\documentclass[sigconf]{acmart}

\AtBeginDocument{%
  \providecommand\BibTeX{{%
\normalfont B\kern-0.5em{\scshape i\kern-0.25em b}\kern-0.8em\TeX}}}

\copyrightyear{2024}
\acmYear{2024}
\setcopyright{acmlicensed}
\acmConference[CIKM '24] {Proceedings of the 33rd ACM International Conference on Information and Knowledge Management}{October 21--25, 2024}{Boise, ID, USA.}
\acmBooktitle{Proceedings of the 33rd ACM International Conference on Information and Knowledge Management (CIKM '24), October 21--25, 2024, Boise, ID, USA}
\acmISBN{979-8-4007-0436-9/24/10}
\acmDOI{10.1145/3627673.3679779}

\usepackage{algorithmic}
\usepackage{graphicx}
\usepackage{xcolor}
\usepackage{colortbl}
\usepackage[linesnumbered,ruled,vlined]{algorithm2e}
\usepackage{multirow}
\usepackage{booktabs}
\usepackage{hyperref} 
\usepackage{subcaption}
\usepackage{url}
\usepackage{utfsym}

\begin{document}

\title{NC$^{2}$D: Novel Class Discovery for Node Classification}


\author{Yue Hou}
\affiliation{%
  \institution{State Key Laboratory of Complex \& Critical Software Environment, Beihang University, Beijing, China}
  \country{}
}
\email{hou\_yue@buaa.edu.cn}

\author{Xueyuan Chen}
\affiliation{%
  \institution{State Key Laboratory of Complex \& Critical Software Environment, Beihang University, Beijing, China}
  \country{}
}
\email{xueyuanchen@buaa.edu.cn}

\author{He Zhu}
\affiliation{%
  \institution{State Key Laboratory of Complex \& Critical Software Environment, Beihang University, Beijing, China}
  \country{}
}
\email{roy_zh@buaa.edu.cn}

\author{Ruomei Liu}
\affiliation{%
  \institution{State Key Laboratory of Complex \& Critical Software Environment, Beihang University, Beijing, China}
  \country{}
}
\email{rmliu@buaa.edu.cn}

\author{Bowen Shi}
\affiliation{%
  \institution{School of Journalism, Communication University of China, Beijing, China}
  \country{}
}
\email{bowenshi@cuc.edu.cn}

\author{Jiaheng Liu}
\affiliation{%
  \institution{State Key Laboratory of Complex \& Critical Software Environment, Beihang University, Beijing, China}
  \country{}
}
\email{liujiaheng@buaa.edu.cn}

\author{Junran Wu}
\authornote{Corresponding author.}
\affiliation{%
  \institution{State Key Laboratory of Complex \& Critical Software Environment, Beihang University, Beijing, China}
  \country{}
}
\email{wu_junran@buaa.edu.cn}

\author{Ke Xu}
\affiliation{%
  \institution{State Key Laboratory of Complex \& Critical Software Environment, Beihang University, Beijing, China}
  \country{}
}
\email{kexu@buaa.edu.cn}

\renewcommand{\shortauthors}{Hou et al.}

\begin{abstract}
Novel Class Discovery (NCD) involves identifying new categories within unlabeled data by utilizing knowledge acquired from previously established categories. However, existing NCD methods often struggle to maintain a balance between the performance of old and new categories. Discovering unlabeled new categories in a class-incremental way is more practical but also more challenging, as it is frequently hindered by either catastrophic forgetting of old categories or an inability to learn new ones. Furthermore, the implementation of NCD on continuously scalable graph-structured data remains an under-explored area. In response to these challenges, we introduce for the first time a more practical NCD scenario for node classification (i.e., NC-NCD), and propose a novel self-training framework with prototype replay and distillation called SWORD, adopted to our NC-NCD setting. Our approach enables the model to cluster unlabeled new category nodes after learning labeled nodes while preserving performance on old categories without reliance on old category nodes. SWORD achieves this by employing a self-training strategy to learn new categories and preventing the forgetting of old categories through the joint use of feature prototypes and knowledge distillation. Extensive experiments on four common benchmarks demonstrate the superiority of SWORD over other state-of-the-art methods.
\end{abstract}






\maketitle

\section{Introduction}
Graph data is commonly used to reveal interactions between various entities, such as academic graphs, social networks, recommendation systems, etc \cite{wu2022simple,wu2022structural,wu2023sega,zhu2023hitin,zhu2024hill}.
In the last few years, node classification has received considerable attention \cite{wang2020gcn, xhonneux2020continuous, wang2021zero, qiu2020gcc, you2021identity} with the rise of significant advances in graph neural networks (GNNs) \cite{kipf2016semi}.
Most of the existing works on node classification primarily focus on a single task, where the model is tasked to classify unlabeled nodes into fixed classes
\cite{hu2019hierarchical, kipf2016semi, oono2019graph, zhuang2018dual}.
In practice, however, many graph data grow continuously, generating new classes as new nodes and edges emerge, for example, in a citation network, the appearance of new papers and their related citations, or even new interdisciplinary; the addition of new users leads to the emergence of new social groups.
The categories of nodes are gradually expanding, usually accompanied by few labels due to their new emergence or lack of exploration.
This requires new category discovery on graphs using GNNs, which currently has limited research on tasks regarding node classification.

\begin{figure*}[!t]
\centering
\subfloat[NC-NCD setting]{
\includegraphics[width=.20\textwidth]{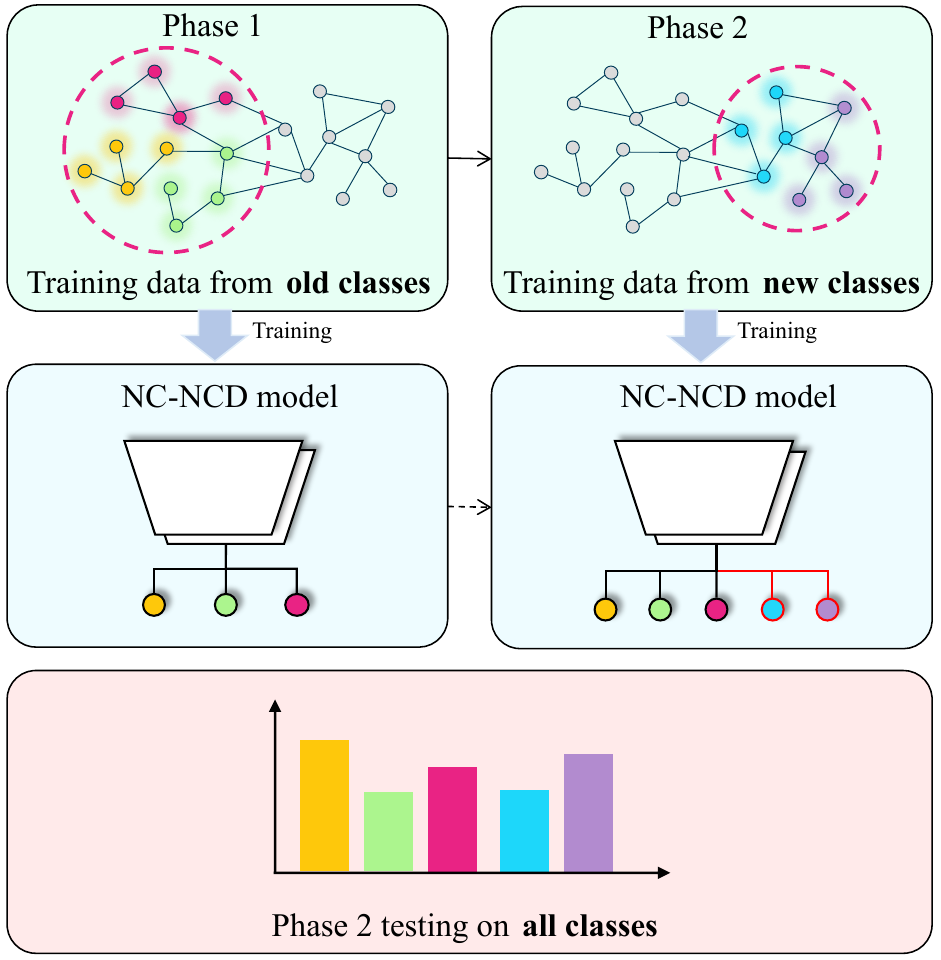}}
\subfloat[NCD setting]{
\includegraphics[width=.20\textwidth]{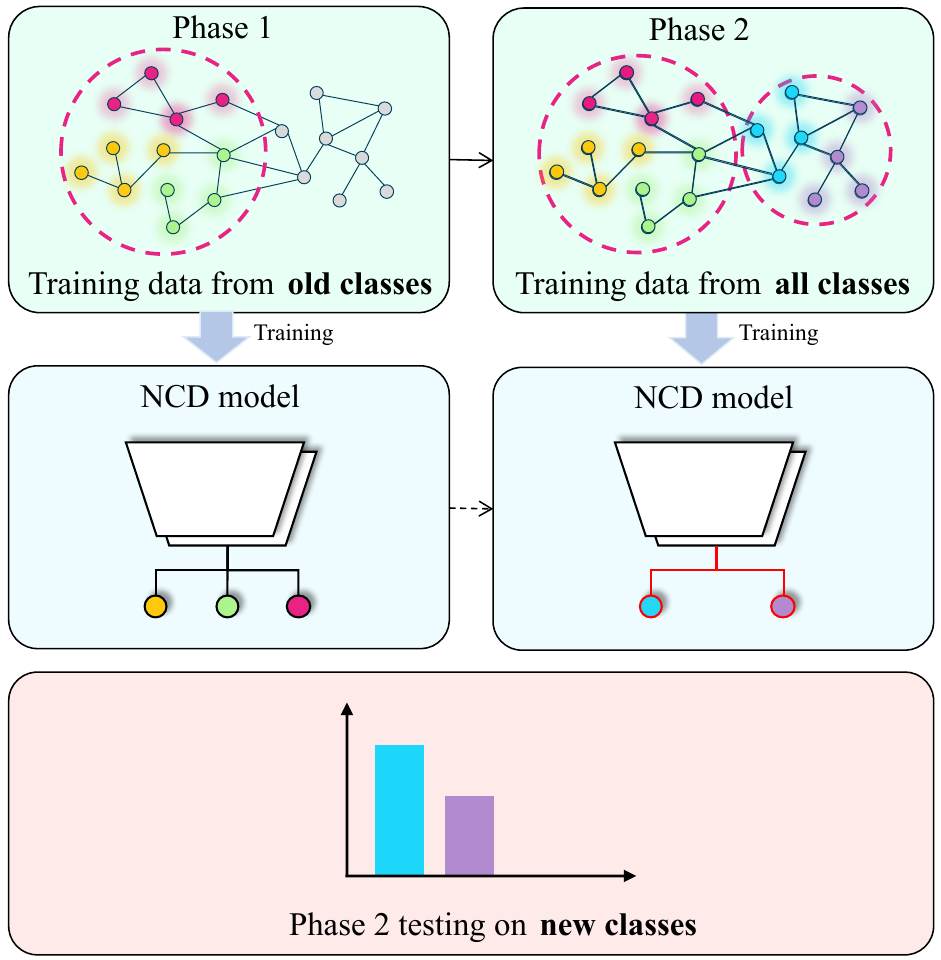}}
\subfloat[NCDwF setting]{
\includegraphics[width=.20\textwidth]{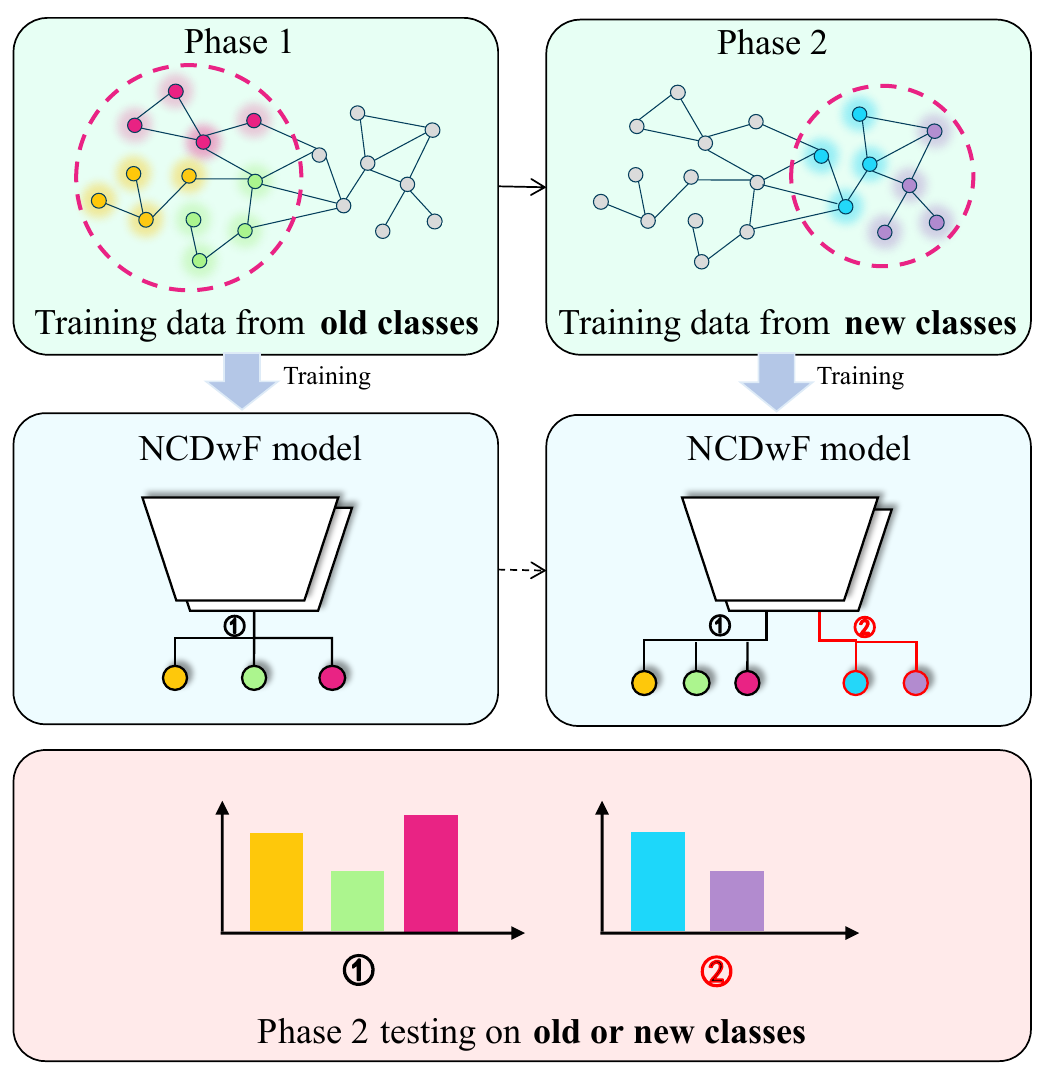}}
\subfloat[task-IL setting]{
\includegraphics[width=.20\textwidth]{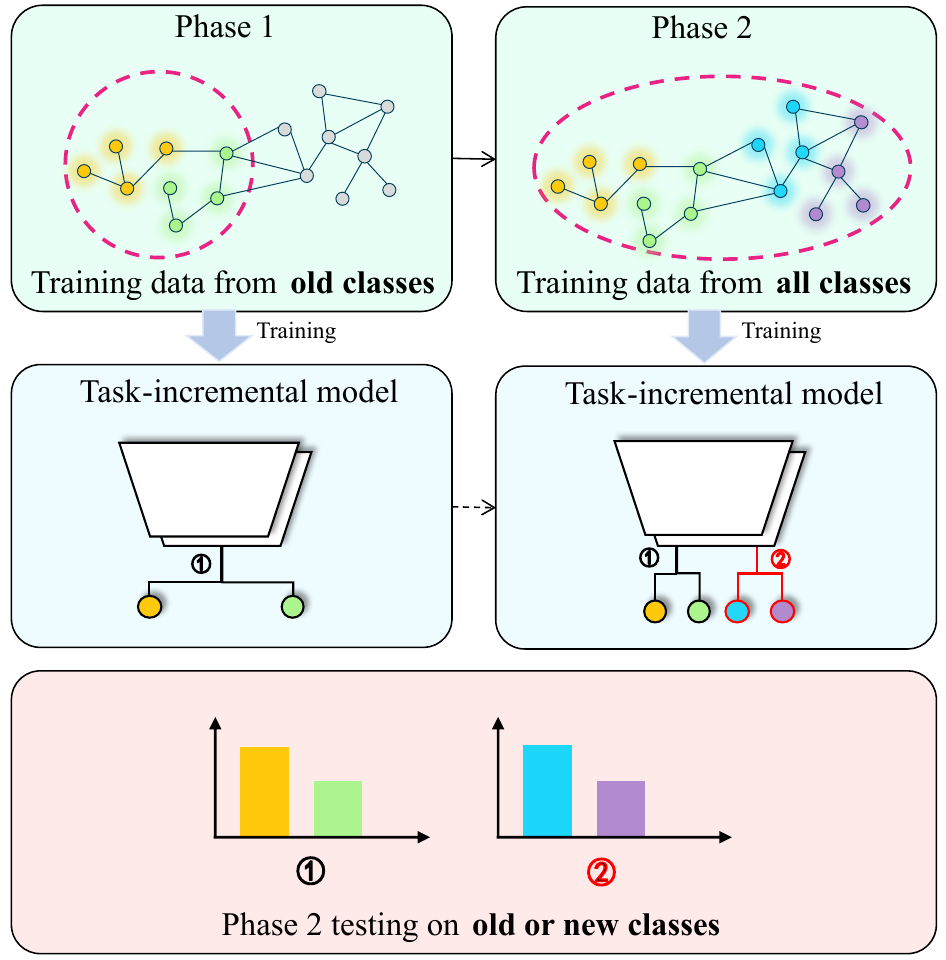}}
\subfloat[class-IL setting]{
\includegraphics[width=.20\textwidth]{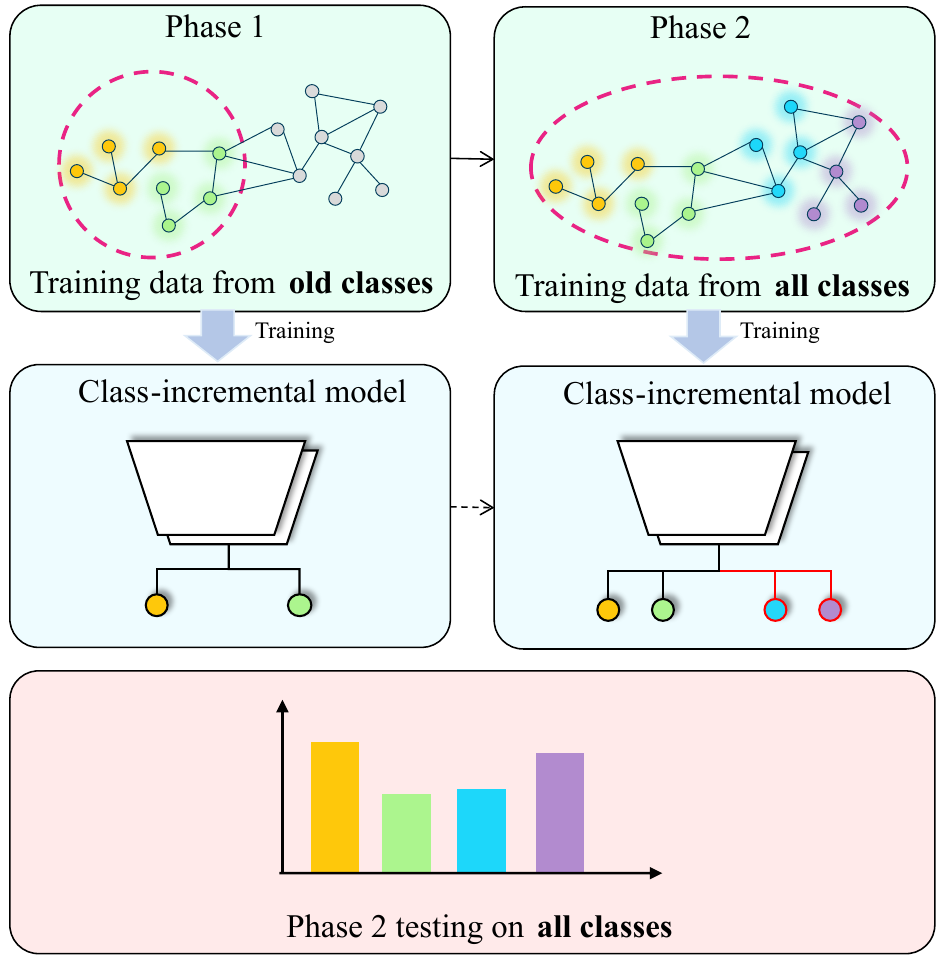}}
  \caption{An illustration of the different settings in novel class discovery and incremental learning tasks. 
  }
  \label{fig:setting}
\end{figure*}

\begin{table*}[!t]
\caption{The difference between task-IL, class-IL, NCD, NCDwF and NC-NCD.}
\label{table:setting}
\centering
\resizebox{\linewidth}{!}{
\begin{tabular}{@{}c|c|cc|c|c|l@{}}
\toprule
\multirow{2}{*}{Settings} & \multirow{2}{*}{Number of classes per task}  & \multicolumn{2}{c|}{Training data} & \multirow{2}{*}{Testing data} & \multirow{2}{*}{Whether task-specific} & \multicolumn{1}{c} {\multirow{2}{*}{Evaluation Protocol}} \\ \cmidrule(lr){3-4}
 &  & \multicolumn{1}{c|}{Phase on learning old classes} & Phase on learning new classes & &  & \multicolumn{1}{c}{} \\ \midrule
task-IL & Same & \multicolumn{1}{c|}{Old classes (labeled)} & Old classes (labeled) + New classes (labeled) & All classes & task-specific & Distinguish old or new classes on each task based on task-ids \\
class-IL & Same & \multicolumn{1}{c|}{Old classes (labeled)} & Old classes (labeled) + New classes (labeled) & All classes & task-agnostic & Distinguish all categories \\
NCD & Not same & \multicolumn{1}{c|}{Old classes (labeled)} & Old classes (labeled) + New classes (unlabeled) & New classes & task-specific & Only distinguish new categories \\
NCDwF & Not same & \multicolumn{1}{c|}{Old classes (labeled)} & New classes (unlabeled) & All classes & task-specific & Distinguish old and new categories with task-specific classifiers \\
\textbf{NC-NCD} & \textbf{Not same} & \multicolumn{1}{c|}{\textbf{Old classes (labeled)}} & \textbf{New classes (unlabeled)} & \textbf{All classes} & \textbf{task-agnostic} & \textbf{Distinguish all categories} \\ 
\bottomrule
\end{tabular}
}
\end{table*}

The task of automatically discovering new classes in an unsupervised manner while utilizing previously acquired knowledge is called novel class discovery (NCD)
\cite{han2020automatically, han2019learning, fini2021unified, zhong2021neighborhood, zhong2021openmix}.
NCD has recently received widespread attention for its ability to effectively learn new classes without relying on large amounts of labeled data.
Most of the currently proposed NCD solutions employ stage-wise \cite{hsu2017learning, hsu2019multi, han2019learning} or
joint \cite{han2020automatically, fini2021unified, zhong2021neighborhood} learning schemes for labeled and unlabeled data.
It has been demonstrated that NCD achieves superior learning performance when clustering unlabeled data and jointly training with labeled data \cite{han2020automatically, zhong2021neighborhood}.
However, joint training relies on the availability of labeled data, which presents two limitations:
(1) labeled data may become unavailable after the pre-training phase due to practical issues such as storage constraints or privacy protection;
(2) retraining labeled data from old classes can make the entire learning process more cumbersome and expensive.
Therefore, a more practical stage-wise NCD setting is needed, where labeled data is discarded and only pre-trained models can be migrated for learning new classes.
Although such a training scheme appears more practical, it gradually leads to the elimination of all previously acquired information about base (or old) classes.
This phenomenon is known as catastrophic forgetting in neural networks \cite{de2021continual}, where the model's performance on previous tasks experiences a significant decline after being trained on a new task.

In light of the above analysis, we propose a novel class discovery task for node classification on graphs called NC-NCD.
The key insight of the NC-NCD setting is to learn a model with a stage-wise scheme that effectively clusters unlabeled nodes from new classes by leveraging knowledge and potential commonalities from labeled nodes belonging to old classes.
Given the inherent limitations of existing NCD settings \cite{liu2022residual}, we argue that the ideal NCD approach is supposed to focus on strictly learning new classes using only unlabeled nodes while simultaneously maintaining performance on the base classes.
Moreover, most of the existing NCD methods mentioned above, whether trained in a stage-wise or joint manner, prioritize learning the current task without considering the global context of all classes. We find these approaches impractical in real-world scenarios, as models adapted to new classes become unusable for base classes and retraining is infeasible.
Inspired by class-IL \cite{tan2022graph}, NC-NCD evaluates the performance on nodes from all classes in the inference phase, thereby overcoming the reliance on the task-id of both old and new classes and truly achieving task-agnostic NCD.

Notice that the NC-NCD setting is essentially different from incremental learning and NCD, and Figure \ref{fig:setting} illustrates the distinctions among these settings.
The goal of NC-NCD is to continue training the model using unlabeled nodes of new categories after learning knowledge from old categories without requiring old category data, which enables the model to classify all categories.
The key distinction between our NC-NCD and the NCD setting lies in the fact that NC-NCD additionally demands the model to maintain performance on old categories, unlike NCD, which focuses solely on the model's performance on new categories. 
Novel Class Discovery without Forgetting (NCDwF) \cite{joseph2022novel} is closely related to NCD but it relaxes certain key assumptions. Notably, existing NCD methods assume access to both labeled and unlabeled data during training, a condition often unrealistic in real-world scenarios. 
Both NCDwF and NC-NCD emphasize the non-concurrent utilization of labeled and unlabeled data during training. However, in contrast to NC-NCD, NCDwF fails to train a classifier capable of predicting across the entire label space. Consequently, the distinction between new and old categories in NCDwF during evaluation still necessitates the utilization of task-ids.
Incremental learning can be categorized into task increment (task-IL) and class increment learning (class-IL), aiming to prevent catastrophic forgetting during the learning of a series of categories that follow a different setting.
Specifically, as depicted in Table \ref{table:setting}, after learning old category nodes, NC-NCD trains on unlabeled nodes of new categories, while task-IL and class-IL require training on both labeled old category and new category nodes. NCD only utilizes new category nodes for training and testing. During the evaluation, task-IL needs to specify the task-id in a task-specific manner, allowing the model to test old or new category nodes with the knowledge of whether the current categories are old or new. In contrast, our NC-NCD distinguishes in a task-agnostic way between all categories seen during evaluation.
Additionally, NC-NCD liberates itself from the constraint of maintaining a consistent number of categories learned in each phase. 
Clearly, our NC-NCD setting is more aligned with real-world application scenarios.

In this work, we propose a novel \textbf{S}elf-training \textbf{W}ith pr\textbf{O}totype \textbf{R}eplay and \textbf{D}istillation framework named \textbf{SWORD}, to implement the NC-NCD task effectively.
Specifically, an encoder for extracting node features is first pre-trained on the labeled old category nodes and used directly in the NCD-training phase.
To facilitate the learning of unlabeled new categories, we set up a classifier specific to the new classes, optimized with robust rank statistics.
Besides, since the NC-NCD setting is task-agnostic, we maintain a joint classifier that is trained with pseudo-labels generated by a task-specific classifier, to classify all the classes encountered throughout the learning process.
Inspired by the effectiveness of replay-based incremental learning \cite{buzzega2020dark, chaudhry2019continual, rebuffi2017icarl},
we replay the prototype of base class nodes after clustering to prevent catastrophic forgetting of base classes. In addition to features encoded from the prototype, feature-level knowledge distillation is employed to uphold the effectiveness of the model's feature extraction.
The main contributions are as follows:

\begin{itemize}
  \item Contrary to existing research in incremental learning and NCD, we introduce a more practical NCD setting for node classification, named NC-NCD.
  \item We propose a novel framework, SWORD, adapted to our NC-NCD setting to address the challenges in the current NCD scenarios formulated with node classification tasks.
  \item Our SWORD trains a task-agnostic joint classifier with a self-training strategy in the NCD-training phase and employs prototype replay and knowledge distillation to prevent forgetting.
  \item Extensive experiments are conducted with multi-backbones to confirm the effectiveness of SWORD by comparing with the state-of-the-art (SOTA) methods on various benchmarks regarding node classification.
\end{itemize}

\section{Related Work}
As information grows, the task of discovering new categories from data and making accurate classification predictions has become a popular research direction \cite{han2020automatically, liu2022residual, zhong2021neighborhood, han2019learning, roy2022class, zhu2024do}.
Here, we devote our attention to NCD for node classification, the most relevant topic of this work.

\vspace{-1.5mm}
\subsection{Graph Representation Learning}
Graph representation learning aims to encode both the feature information from nodes and the topological structure of the incoming graph.
Inspired by word2vec \cite{mikolov2013distributed}, several methods, such as DeepWalk \cite{perozzi2014deepwalk} and node2vec \cite{grover2016node2vec}, generate sequences of nodes by random walks, and apply skip-gram approach to map nodes within the same context to similar vector representations. Although these methods show success in various tasks  like node classification and link prediction, they overemphasize proximity information at the expense of structural information and fail to incorporate node attributes.

Recently, GNNs have received a great deal of attention,
such as graph convolutional networks (GCNs) \cite{kipf2016semi}, GraphSAGE \cite{hamilton2017inductive}, graph attention networks (GATs) \cite{velivckovic2017graph}
and their extensions \cite{xu2018powerful, chen2018fastgcn, bahonar2019graph, zhang2019graph, zhang2020context, lei2020spherical, gao2021topology}.
GNNs take in high-dimensional node features and map them to low-dimensional vector embeddings that are used for different downstream tasks.
Based on methods of aggregation, GNNs are mainly classified into spectral and spatial embedding.
The spectral approaches \cite{defferrard2016convolutional, kipf2016semi} learn graph representations by the Fourier transform of graph data, while spatial approaches perform simple spatial aggregation of features from neighbors \cite{hamilton2017inductive, velivckovic2017graph}.
However, most of these works are designed for a single learning task and are not directly applicable to incremental learning problems, thus encountering catastrophic forgetting problems when learning a series of new tasks.

\vspace{-1.5mm}
\subsection{Novel Class Discovery}
NCD task is to migrate knowledge learned from labeled datasets to unlabelled ones to discover new classes, assuming that the classes in labeled and unlabelled sets are disjoint.
The proposed NCD methods can be broadly divided into two categories.
The first category of approach utilizes a joint training scheme, assuming the availability of both labeled and unlabeled data \cite{wang2020open, yoon2020vime, yu2022self}.
The other category employs a stage-wise isolated training scheme, where data from old and new classes cannot be used simultaneously.
In this context, the labeled data of the base classes are used only during the supervised pre-training phase and not thereafter.
Instead, an unsupervised clustering loss is employed to fine-tune unlabelled data during the phase of discovering unknown new classes \cite{hsu2019multi, yang2010image, zhong2021neighborhood}.

According to \cite{liu2022residual}, NCD methods that rely on joint training always outperform methods with isolated training.
However, it is not always practical to have access to labeled data after the pre-training phase due to concerns about data privacy or storage limitations. Therefore, training on the old and new classes jointly may not be a viable option in many cases.
Moreover, earlier approaches based on isolated training schemes fail to adequately address the issue of catastrophic forgetting, which can result in models losing their ability to classify base classes.
To tackle this problem, a novel task called Novel Class Discovery without Forgetting (NCDwF) is proposed in \cite{joseph2022novel}, which presents a more challenging domain derived from NCD.
ResTune \cite{liu2022residual} and FRoST \cite{roy2022class} are the first isolated training methods to address NCDwF. ResTune incorporates the use of the Hungarian Assignment \cite{kuhn1955hungarian} that concatenates classifiers specific to the old and new classes. However, despite these efforts, ResTune still incorrectly evaluates performance due to confusion between the old and new classes.

In this work, since the implementation of NCD tasks on the graph is almost non-existent, we migrate NCD to the node classification task with GNN and propose a more reasonable NC-NCD setting.
Specifically, the assumption of NC-NCD differs from that of the current NCD, primarily manifested in the inference phase of performance evaluation for classifications, employing a task-agnostic joint classifier.

\begin{figure*}[!t]
\vspace{-0.2cm}
\centering
\includegraphics[width=\linewidth]{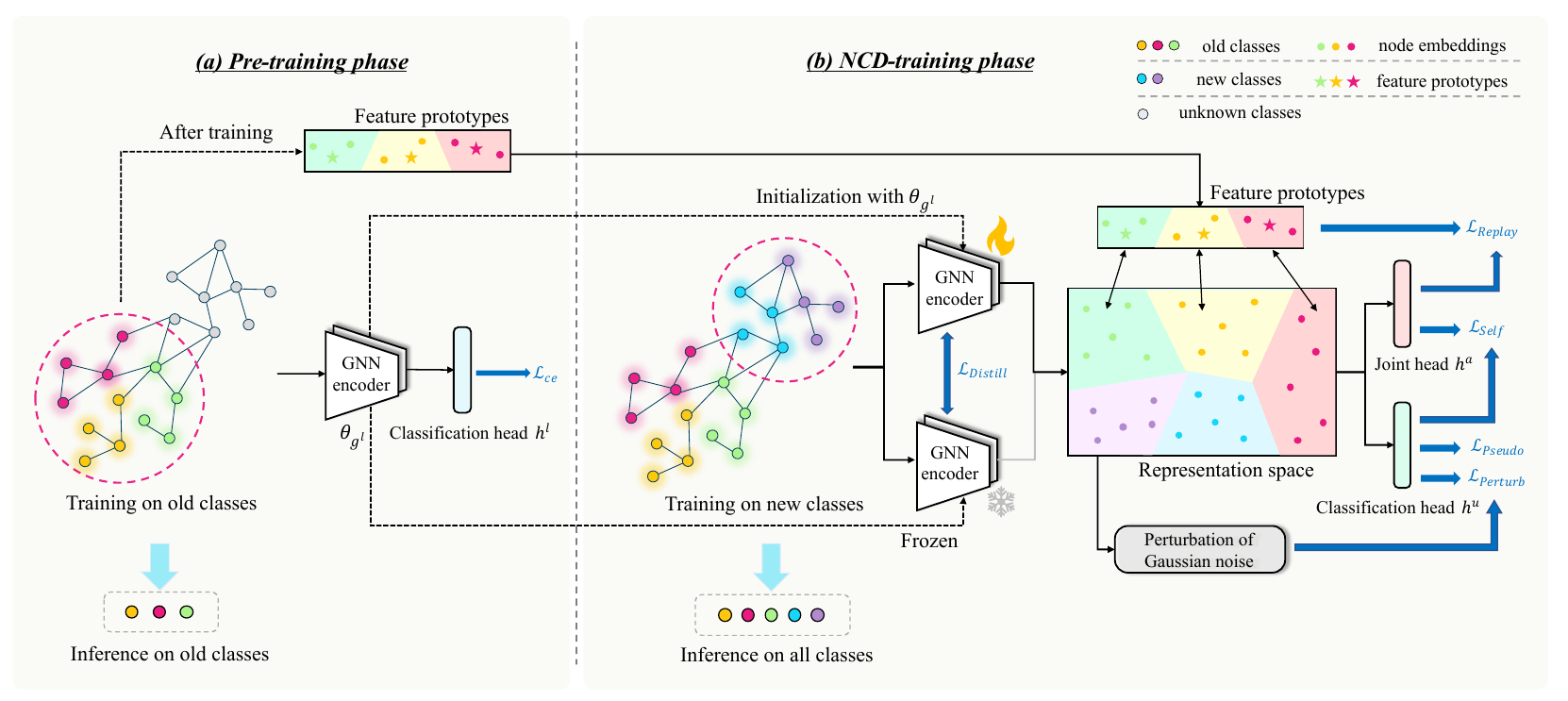}
\caption{Overall architecture of SWORD.
(a) Pre-training phase, a GNN encoder that extracts node representations is trained on labeled old classes and feature prototypes are recorded for the NCD task;
(b) NCD-training phase, SWORD learns unlabeled new category nodes and prevents forgetting through self-training with prototype replay and distillation.
}
\label{fig:sword}
\vspace{-0.2cm}
\end{figure*}

\vspace{-1.5mm}
\subsection{Incremental Learning}
In the past, incremental learning (i.e., continual learning or lifelong learning) has been explored in several areas
such as computer vision 
\cite{hayes2020lifelong, wu2021striking},
and reinforcement learning 
\cite{rolnick2019experience}.
Essentially, it aims to address the problem of catastrophic forgetting while being flexible enough to continue learning on new tasks.

Incremental learning can be divided into task-incremental (task-IL) and class-incremental (class-IL) settings, based on whether task indicators are provided to the model during testing:
\begin{itemize}
  \item task-IL: the model needs to reason about the specified task using task indicators, and the model only needs to distinguish between classes within the current task without considering classes from pre-existing tasks.
  \item class-IL: no task indicators are provided, and the model must differentiate between all classes in both current and pre-existing tasks.
\end{itemize}

While conventional incremental learning methods are abundant, there have been limited efforts to apply incremental learning techniques to graphs, i.e., graph incremental learning (a.k.a. continual graph learning) \cite{wang2022lifelong, xu2020graphsail, daruna2021continual, kou2020disentangle, ahrabian2021structure, cai2022multimodal}.
Existing approaches on graph incremental learning can be categorized into three types, i.e., 
memory replay methods (e.g., Experience Replay Graph Neural Network (ER-GNN) \cite{zhou2021overcoming}) which stores representative nodes which are replayed when learning new tasks; 
parameter isolation methods (e.g., Hierarchical Prototype Networks (HPNs) \cite{zhang2022hierarchical}) which adaptively select different parameter combinations for different tasks; 
and regularization based methods (e.g., Topology-aware Weight Preserving (TWP) \cite{liu2021overcoming}) which preserves crucial parameters and topologies via regularization. 

For graph incremental learning tasks regarding node classification, such as the common application scenarios of multi-class classification like citation networks or social networks,
it is evident that the class-IL setting is more practical and challenging.
However, existing graph incremental learning methods mentioned earlier are limited to a specific set of tasks with labeled data, which restricts its practicality and does not align with our setting.

In this paper, our proposed SWORD is the first attempt in the NC-NCD setting on graphs,
to effectively mitigate the forgetting of old categories while learning new categories when labeled data are no longer available.

\section{Methodology}
In this section, \textbf{SWORD}~\footnote{The code of SWORD is available at \url{https://github.com/name-is-what/NC2D}.} are elaborated below.
First, several preliminary definitions and notations are given.

\subsection{Novel Class Discovery for Node Classification}
\subsubsection{Problem Definition and Notation}
Formally, a graph $\mathcal{G}=\{\mathcal{V}, \mathcal{E}\}$, where $\mathcal{V}$ is the set of $\mathrm{N}$ nodes and $\mathcal{E}$ is the edge set, can be alternatively represented by $\mathcal{G}=(\mathrm{A}, \mathrm{X})$.
Let $\mathrm{A} \in \mathbb{R}^{\mathrm{N} \times \mathrm{N}} $ be the adjacency matrix and $\mathrm{X} \in \mathbb{R}^{\mathrm{N} \times d_0}$ denotes the node attribute matrix, where $d_0$ denotes the dimension of node attributes.
To solve the NC-NCD problem, we split the dataset $\mathcal{G}$ into two folds with disjoint categories, $\mathcal{G}^l$ and $\mathcal{G}^u$ to represent data from the base and new classes respectively.
$\mathcal{G}^l$ is randomly split into $\mathcal{G}^l_{train}$, $\mathcal{G}^l_{val}$, and $\mathcal{G}^l_{test}$ for pre-training phase.
$\mathcal{G}^u$ also has three splits of $\mathcal{G}^u_{train}$ , $\mathcal{G}^u_{val}$ , $\mathcal{G}^u_{test}$.
To simplify the writing, we uniformly use $\mathcal{G}^l$, $\mathcal{G}^u$, $\mathcal{G}^a$ to denote data from base, new and all classes.

For supervised task $\mathcal{T}^l$, given labeled $\mathcal{G}^{l} = \{(\mathrm{A}, \mathrm{x}_i^{l}, {y}_i^{l} )\}_{i=1}^{\mathrm{N}^{l}}$ with $\mathrm{N}^l$ node samples, where $\mathrm{x}_i^{l} \in \mathrm{X}^{l}$ represents the feature of the current node and ${y}_i^{l} \in \mathcal{Y}^{l}$ is $\left\|\mathcal{C}^l\right\|$-dimensional one-hot labels.
Once the task $\mathcal{T}^l$ is completed, $\mathcal{G}^{l}$ is discarded and we get $n^{u}$ instances from the self-supervised new task $\mathcal{T}^u$.
Task $\mathcal{T}^u$ employs an unlabeled  $\mathcal{G}^{u} = \{(\mathrm{A}, \mathrm{x}_i^{u})\}_{i=1}^{\mathrm{N}^{u}}$ , where $\mathrm{x}_i^{u} \in \mathrm{X}^{u}$ is the unlabeled nodes from class $\mathcal{C}^u$.
The number of novel classes $\left\|\mathcal{C}^u\right\|$ is generally treated as a prior \cite{han2020automatically, liu2022residual}, as it can often be effectively estimated through methods such as semi-supervised k-means.
Like other NCD settings \cite{han2019learning}, the labels in $\mathcal{Y}^{l}$ and $\mathcal{Y}^{u}$ are assumed to be disjoint, i.e., $\mathcal{Y}^{l} \cap \mathcal{Y}^{u} = {\varnothing} $.
The goal of NC-NCD is to cluster nodes in $\mathcal{G}^u$ on task $\mathcal{T}^u$ while still performing well on $\mathcal{T}^l$ by exploiting the encoding ability in the mapping function $f^{l}:\mathcal{G}^l \rightarrow \mathcal{Y}^l$. In a nutshell, we learn a mapping function $f:\mathcal{G}^u\rightarrow \mathcal{Y}^l\cup \mathcal{Y}^u$ on $\mathcal{T}^u$ that can infer all classes of nodes from $\mathrm{X}^l\cup  \mathrm{X}^u$, which is the major difference from existing NCD methods.

\vspace{-1.5mm}
\subsubsection{Evaluation Protocol}
Previous NCD methods \cite{fini2021unified, zhong2021neighborhood} trained task-specific classification heads for the base and new classes respectively.
To overcome the limitation of NCD being restricted to specified tasks, ResTune \cite{liu2022residual} introduced the use of Hungarian Assignment \cite{kuhn1955hungarian} to concatenate these two classification heads to achieve an evaluation of all classes during inference.
However, the Hungarian Assignment which is a direct maximum match of the outputs from the two heads, lacks the ability to distinguish between old and new classes when they become completely confused.

In the present work, our NC-NCD setting uses a more reasonable evaluation protocol. Specifically, after learning the labeled nodes in base classes, since the unlabeled data in the NCD stage is used for isolated training, we employ a classification head for new classes to assign labels specifically and use pseudo-labels to train a single joint classifier that can classify both the base and new classes. When evaluating the model, the trained task-agnostic joint classifier is used directly without specifying the task-id, and the classification results are inferred on all seen class data, compared with the ground-truth labels of the samples. It is clear that our evaluation protocol is more reasonable, such as being able to discern when a new category is classified as one of the old categories is an error, which is a desirable behavior.

\vspace{-1.5mm}
\subsubsection{Overall Framework}
Within NC-NCD setting, the function $f$ consists of two major components: a GNN encoder $g(\cdot)$ to learn node representations and a linear classification head $h(\cdot)$ to output logits.
Our proposed SWORD runs in two phases, a pre-training phase on the supervised task $\mathcal{T}^l$ and an NCD-training phase on the unsupervised task $\mathcal{T}^u$.

\vspace{-1.5mm}
\paragraph{Pre-training}
In this stage, we utilize the labeled nodes from $\mathcal{G}^l$ to learn the mapping function $f^l = h^l \circ g$ in a supervised manner, which identifies samples belonging to the category $\mathcal{C}^l$. The feature extractor $g$ and the classifier $h^l$ are parameterized by $\theta_{g^l}$ and $\theta_{h^l}$, respectively.
The latent features of the old category nodes are defined as $\mathrm{z}^l = g(\mathcal{G}^l)$.
We aim to learn these two parameters $\{\theta_{g^l}, \theta_{h^l}\}$ of model $f^l$ by using supervised cross-entropy loss:
\begin{equation}
  \label{equa:loss_ce_label}
  \mathcal{L}_{ce}=-\frac{1}{ \left\|\mathcal{C}^l\right\|} \sum_{k=1}^{  \left\|\mathcal{C}^l\right\|} y_{k}^l\log \sigma_{k}\left( h^l ( \mathrm{z}^l )\right),
\end{equation}
where the softmax function $\sigma_{k}=\frac{\exp ({l}_k)}{\sum_{j} \exp({l}_k)}$ represents the likelihood corresponding to the $k^{th}$ output from the model and $\left\|\mathcal{C}^l\right\|$ is the number of classes in the task $\mathcal{T}^l$.
In addition, before the next stage, we will compute the feature prototype $\boldsymbol{\mu_c}$ and variance $\boldsymbol{{\nu_c}^2}$ with feature $\mathrm{z}^l$ for each class $c$ in $\mathcal{G}^l$.

\vspace{-1.5mm}
\paragraph{NCD-training}
In the second stage, we learn new categories on the unlabeled data $\mathcal{G}^u$ by reusing the parameters from the pre-trained model $f^l$. In the NC-NCD setting, the model $f$ should have a joint classifier to accommodate $\mathcal{C} =\mathcal{C}^l + \mathcal{C}^u$ categories. Therefore, we directly extend the classifier $h^l$ into $h^a$ and provide a specific classifier $h^u$ separately for the new category's classification on the $\mathcal{T}^u$ task. These two classifiers are parameterized by $\{\theta_{h^a}, \theta_{h^u}\}$.
To be specific, we train $f^u=h^u \circ g$, where $\mathcal{G}^u \rightarrow \mathcal{Y}^u$, through the SWORD framework using a clustering objective that provides supervision on previously learned information using robust rank statistics and prevents forgetting of old categories through prototype replay and distillation. Function $f =h^a \circ g$, where $\mathcal{G} \rightarrow \mathcal{Y}^l\cup \mathcal{Y}^u$, is used during inference. More details are given in the next section.

\subsection{Self-training with Prototype Replay and Distillation}
We aim to learn a model that has a better capability of clustering unlabeled nodes after pre-training with labeled data while preserving the performance on previously seen categories without visiting nor storing previously labeled nodes.
Our SWORD framework will achieve the above task requirements through self-training, prototype replay, and distillation, which are explained in detail below.

\vspace{-1.5mm}
\subsubsection{Pairwise Similarity and Pseudo Label}
In the NCD-training phase, an essential step in learning new classes is to train $f^u=h^u \circ g$. On the one hand, we update the weights of GNN encoder $g$ to maintain the feature extraction capability on new category nodes. On the other hand, we train the task-specific classifier $h^u$, which is initialized and lacks the ability to discriminate the features on unlabeled nodes.
We adopt the ideology from the NCD method AutoNovel \cite{han2020automatically} to infer pairwise similarity between features of a pair of unlabeled nodes and employ it in a weakly supervised form during NCD-training steps.
Specifically, given a pair of node-featured pairs $(\mathrm{x}^u_i, \mathrm{x}^u_j)$ taken from the unlabeled $\mathcal{G}^u$, we use GNN encoder $g$ to compute $\mathrm{z}^u_i = g(\mathcal{G}^u_i)$ and $\mathrm{z}^u_j = g(\mathcal{G}^u_j)$, respectively, to obtain the pair of node representations $(\mathrm{z}^u_i, \mathrm{z}^u_j)$ on the graph.
Then we multiply the node representation pair between the two samples with dot product and normalize it using a logistic function $\sigma(\cdot)$. That is, we calculate the pairwise similarity using the following:
\begin{equation}
  \label{equa:pair_simi}
  s_{ij} = \sigma (h^u(\mathrm{z}^u_i)\odot h^u(\mathrm{z}^u_j)).
\end{equation}
Since the class of nodes in this stage is unknown, robust rank statistics are utilized to compare the node representation pairs. We consider that if the top-k dimensions of node representation pair $(\mathrm{z}^u_i, \mathrm{z}^u_j)$ are the same, the two nodes corresponding to index $i$ and $j$ are considered to belong to the same class. In this way, another pairwise similarity of the two sample nodes is generated, referred to as a pairwise pseudo label to weakly supervise the training of classification head $h^u$ for new classes.
The pairwise pseudo labels are formulated as:
\begin{equation}
  \label{equa:pair_pseudo}
  \tilde{y}_{ij}^u =\mathbf{1}\left\{ \mathrm{top_k}(\mathrm{z}^u_i) = \mathrm{top_k}(\mathrm{z}^u_j) \right\},
\end{equation}
where $\mathrm{top_k}: \mathrm{z}^u \rightarrow \mathcal{S} \left\{ (1, \ldots, \mathrm{k})\right\} \subset \mathcal{P}\left\{ (1, \ldots,\left\| \mathrm{z}^u \right\|)\right\}$.
In particular, ranking the values by size in node representation vectors, if $\mathrm{z}^u_i$ and $\mathrm{z}^u_j$ are identical in the top-k most activated dimensions, then pairwise pseudo label $\tilde{y}_{ij}^u = 1$ for nodes $i$ and $j$, and 0 otherwise.
Based on the pairwise pseudo label $\tilde{y}_{ij}^u$ and pairwise similarity $s_{ij}$, the pairwise labeling objective with parameters $\{\theta_{g}, \theta_{h^u}\}$ is trained with binary cross-entropy loss as:
\begin{equation}
  \label{equa:loss_pseudo}
  \mathcal{L}_{Pseudo}= -\frac{1}{|\mathrm{N}^u|^{2}} \sum_{i,j}^{ |\mathrm{N}^u|} \left(
  \tilde{y}_{ij}^u \log s_{ij}+(1-\tilde{y}_{ij}^u) \log (1-s_{ij}) \right).
\end{equation}
This equation trains a task-specific classifier for the learning of new category nodes, which allows the model to cluster unlabeled nodes on task $\mathcal{T}^u$ for the next NCD-training steps.

\vspace{-1.5mm}
\subsubsection{Self-training with Joint Head}
To accommodate the NC-NCD setting, the inference step of our training framework should ideally not depend on task-id.
We thus train the joint classifier $h^a$ by self-training with the help of pseudo labels computed from $f^u=h^u \circ g$.
In summary, given the goal of the learning model $f$, we utilize the task-specific classification head $h^u$ to assign the pseudo label $\hat{y}^u$ to the unlabeled nodes and use this pseudo label to supervise the training of the joint classification head $h^a$, where the pseudo label $\hat{y}^u$ of unlabeled nodes can be computed as:
\begin{equation}
  \label{equa:pseudo_label}
  \hat{y}^u =\left\|\mathcal{C}^l\right\| +\underset{k \in \left\|\mathcal{C}^u\right\|}{\arg \max } h^u(\mathrm{z}^u).
\end{equation}
The self-training loss is described as:
\begin{equation}
  \label{equa:loss_self}
  \mathcal{L}_{Self}=-\frac{1}{\left\|\mathcal{C}^a\right\|} \sum_{k=1}^{\left\|\mathcal{C}^a\right\|} \hat{y}_{k}^u \log \sigma_{k}\left(h^a(\mathrm{z}^u)\right).
\end{equation}

\vspace{-1.5mm}
\subsubsection{Encoder Output Perturbation}
Since the pairwise pseudo labels obtained in Eq.(\ref{equa:pair_pseudo}) can be noisy, it will lead to poor supervised training of $h^u$. Then it will affect the pseudo labels in Eq.(\ref{equa:pseudo_label}), which adversely affects the training of the joint classifier $h^a$. Therefore, to minimize the impact of the noise cascade propagation from $h^u$, inspired by stochastic data augmentation, instead of changing the structural information of the graph itself, we add perturbations to encoder output features on node representation level \cite{xia2022simgrace} to perturb classifier $h^u$.
Specifically, we directly add the perturbation of random Gaussian noise to the node representation $\mathrm{z}^u$ obtained with GNN encoder, described mathematically as,
\begin{equation}
  \label{equa:z_perturb}
  \mathrm{z}_k^{\prime}=\mathrm{z}_k^u + \eta \cdot \Delta \mathrm{z}_k; \quad \Delta \mathrm{z}_k \sim \mathcal{N}\left(0, \sigma_k^2\right),
\end{equation}
where $\eta$ is the coefficient that scales the magnitude of the perturbation, and $ \Delta \mathrm{z}_k$ is the perturbation term which samples from Gaussian distribution with zero mean and variance $\sigma_k^2$.
The $h^u$ is reused for representation with perturbation $\mathrm{z}^{\prime}$ and original $\mathrm{z}^u$, with a mean-squared error loss to further optimize the parameters $\{\theta_{g}, \theta_{h^u}\}$:
\begin{equation}
  \label{equa:loss_perturb}
  \mathcal{L}_{Perturb} = \frac{1}{\left\|\mathcal{C}^u\right\|} \sum_{k=1}^{\left\|\mathcal{C}^u\right\|} \left [ \sigma_{k}\left(h^a(\mathrm{z}^u)\right) - \sigma_{k}\left(h^a(\mathrm{z}^{\prime})\right) \right ]^{2}.
\end{equation}

\vspace{-1.5mm}
\subsubsection{Prototype Replay and Distillation}

Although the self-training described above helps the model $f =h^a \circ g$ discover new categories, at the same time it loses the ability to predict old categories on task $\mathcal{T}^u$. So, we propose feature-level prototype replay and distillation.
In the NC-NCD setting, with labeled data $\mathcal{G}^l$ on task $\mathcal{T}^u$ that has been discarded, we use the node representation $\mathrm{z}^l$ at the end of supervised training for task $\mathcal{T}^l$ to compute and record the prototype $\boldsymbol{\mu_c}$ and the variance $\boldsymbol{\nu_c^2}$ in advance for each base class as:

\begin{equation}
  \label{equa:mu_nu}
  \boldsymbol{\mu_c} = \frac{1}{|\mathrm{N}_c^l|} \sum_{i=1}^{|\mathrm{N}_c^l|} g^l(\mathcal{G}^l_i),
  \quad \boldsymbol{\nu_c^2} = \frac{1}{|\mathrm{N}_c^l|} \sum_{i=1}^{|\mathrm{N}_c^l|}\left[g^l(\mathcal{G}^l_i)-\boldsymbol{\mu}_c\right]^2,
\end{equation}
where the number of samples belonging to class $c$ in labeled data $\mathcal{G}^l$ is $|\mathrm{N}_c^l|$.

In the task $\mathcal{T}^u$ of learning new nodes, we specifically replay these prototype features from the class-specific Gaussian distribution $\mathcal{N} \left(\boldsymbol{\mu}_c, \boldsymbol{v}_c^2\right)$, in order to update the parameter $\theta_{h^a}$ and maintain the performance on base classes. The prototype replay loss is given as:

\begin{equation}
  \label{equa:loss_replay}
  \mathcal{L}_{Replay} = -\frac{1}{ \left\|\mathcal{C}\right\|} \sum_{k=1}^{  \left\|\mathcal{C}\right\|} y_{kc}^l \log \sigma_{k}\left( h^a ( \mathrm{z}_c^l )\right),
\end{equation}
where $\left(\mathrm{z}_c^l, {y}_c^l \right) \sim \mathcal{N} \left(\boldsymbol{\mu}_c, \boldsymbol{v}_c^2\right)$.

In addition, the parameters of the feature extractor $g$ are optimized and updated during the self-training step, and the encoder output perturbation only updates the parameters $\theta_{g}$ based on new category nodes, which can cause the prototype of the base class to become obsolete.
To prevent the encoder from losing its effectiveness for prototype feature extraction, we add an additional regularization to $g$ by feature distillation, given as follows:

\begin{equation}
  \label{equa:loss_kd}
  \mathcal{L}_{Distill} = -\frac{1}{|\mathrm{N}^u|} \sum_{i}^{ |\mathrm{N}^u|}
\left\|g^l(\mathcal{G}^u_i) - g(\mathcal{G}^u_i) \right\|_2,
\end{equation}
where $g^l$ is the feature extractor from the previous task and is kept frozen with parameters $\theta_{g^l}$.
Our objective is to encourage the alignment of the feature extractors, denoted as $g$ and $g^l$, with the distribution of new category data, minimizing their differences. 
This alignment facilitates the knowledge distillation from the old model to the new model.

\vspace{-1.5mm}
\subsection{Overall Training}
As aforementioned, the above five loss functions can be mainly divided into two types, one is the novel classes self-training loss to ensure the discovery and clustering of unlabeled nodes in the NCD-training stage, and the other is the base classes replaying loss to prevent the forgetting of the old class nodes in the previous task loss.
Novel classes self-training loss can be written as:
\begin{equation}
  \label{equa:loss_novel}
  \mathcal{L}_{Novel} = \mathcal{L}_{Pseudo} + \beta_1 \mathcal{L}_{Self} + \beta_2 \mathcal{L}_{Perturb},
\end{equation}
where $\beta_1$ and $\beta_2$ are ramp-up functions to ensure stability in learning.
Base classes replaying loss can be formulated as:
\begin{equation}
  \label{equa:loss_base}
  \mathcal{L}_{Base} = \mathcal{L}_{Replay} + \omega_{fd} \mathcal{L}_{Distill},
\end{equation}
where $\omega_{fd}$ is a coefficient set to 10 by default from FRoST \cite{roy2022class}.
In addition, the parameter $\lambda$ is introduced to provide the model with the ability to better balance the performance of base and new classes.
The overall loss is defined as the combination of the novel classes self-training loss and the base classes replaying loss:
\begin{equation}
  \label{equa:loss_overall}
  \mathcal{L} = \mathcal{L}_{Novel} + \lambda \mathcal{L}_{Base}.
\end{equation}
The overall process of SWORD is shown in Algorithm \ref{algo:overall} in Appendix~\ref{app:algo}.

\begin{table*}[!t]
\caption{A summary and comparison of different methods in the fields of NCD, task-IL, class-IL and NC-NCD.}
\label{table:baseline-setting}
\centering
\resizebox{0.9\linewidth}{!}{
\begin{tabular}{@{}c|c|c|c|c|c@{}}
\toprule
\multirow{2}{*}{Methods} & \multirow{2}{*}{Settings} & \multicolumn{2}{c|}{Training on Phase 2} & \multirow{2}{*}{Need task-ids} & \multirow{2}{*}{Evaluation Protocol} \\ \cmidrule(lr){3-4}
 &  & \multicolumn{1}{c|}{With Old Classes} & {New Classes Labeled} &  & \multicolumn{1}{c}{} \\ \midrule

AutoNovel \cite{han2020automatically} & NCD & $\checkmark$ & $\usym{2715}$ & $\usym{2715}$ & Only distinguish new classes \\ 
NCL \cite{zhong2021neighborhood} & NCD & $\checkmark$ & $\usym{2715}$ & $\usym{2715}$ & Only distinguish new classes \\ 
DTC \cite{han2019learning} & NCD & $\checkmark$ & $\usym{2715}$ & $\usym{2715}$ & Only distinguish new classes \\ 
ResTune \cite{liu2022residual} & NCDwF & $\usym{2715}$ & $\usym{2715}$ & $\checkmark$ & Distinguish all categories \\ 
GEM \cite{lopez2017gradient} & task-IL & $\checkmark$ & $\checkmark$ & $\checkmark$ & Distinguish old or new classes on each task\\ 
ER-GNN \cite{zhou2021overcoming} & task-IL & $\checkmark$ & $\checkmark$ & $\checkmark$ & Distinguish old or new classes on each task\\
TWP \cite{liu2021overcoming} & task-IL & $\checkmark$ & $\checkmark$ & $\checkmark$ & Distinguish old or new classes on each task\\ 
CPCA \cite{ren2023incremental} & class-IL & $\checkmark$ & $\checkmark$ & $\usym{2715}$ & Distinguish all categories \\ 
\textbf{SWORD} & \textbf{NC-NCD} & $\usym{2715}$ & $\usym{2715}$ & $\usym{2715}$ & \textbf{Distinguish all categories} \\ \bottomrule
\end{tabular}
}
\end{table*}

\section{Experiment}
\subsection{Experimental Setups}

\paragraph{Dataset Settings}
We evaluate the proposed SWORD on four representative datasets: Cora \cite{mccallum2000automating}, Citeseer \cite{giles1998citeseer}, Pubmed \cite{sen2008collective} and Wiki-CS \cite{mernyei2020wiki}, whose details are provided in Appendix~\ref{app:datasets}.
We also provide statistics and splitting for the datasets in Table \ref{table:dataset}.

\vspace{-1.5mm}
\paragraph{Baselines}
We choose representative GNNs as backbones including GCN \cite{kipf2016semi}, GAT \cite{velivckovic2017graph} and GraphSAGE \cite{hamilton2017inductive}.
Additionally, we compare our SWORD with several SOTA baselines, including 4 traditional NCD methods (AutoNovel \cite{han2020automatically}, ResTune \cite{liu2022residual}, NCL \cite{zhong2021neighborhood} and DTC \cite{han2019learning}), and 4 graph incremental learning methods in both task-IL and class-IL settings (GEM \cite{lopez2017gradient}, ER-GNN \cite{zhou2021overcoming}, TWP \cite{liu2021overcoming} and CPCA \cite{ren2023incremental}).
Table \ref{table:baseline-setting} provides a further summary and comparison of these baseline methods with our SWORD.
A more detailed analysis of these baseline methods is described in Appendix~\ref{app:baselines}.

We also provide implementation details and parameter settings in Appendix~\ref{app:Implementation details}.

\begin{table*}[!t]
\caption{Comparison with SOTA methods on the results of node classification accuracy in NC-NCD setting.
Old, New, and ALL represent the evaluation of the model's classification performance on old category nodes, new category nodes, and nodes of all previously seen categories, respectively.
The numbers in brackets represent the number of corresponding categories.
}

\label{table:comp}
\centering
\resizebox{\linewidth}{!}{
\begin{tabular}{@{}c|c|ccc|ccc|ccc|ccc|ccc@{}}
\toprule
\multirow{2}{*}{Backbone} & \multirow{2}{*}{Methods} & \multicolumn{3}{c|}{Cora} & \multicolumn{3}{c|}{Citeseer} & \multicolumn{3}{c|}{Pubmed} & \multicolumn{3}{c|}{Wiki-CS} & \multicolumn{3}{c}{Average} \\ \cmidrule(l){3-17}
  &  & Old(4) & New(3) & \textbf{All}(7) & Old(4) & New(2) & \textbf{All}(6) & Old(2) & New(1) & \textbf{All}(3) & Old(7) & New(3) & \textbf{All}(10) & Old & New & \textbf{All} \\ \midrule

\multirow{8}{*}{GCN} & AutoNovel & 19.01 & 0.00 & 13.00 & 11.48 & 0.00 & 7.70 & 69.65 & 0.00 & 41.30 & 30.09 & 0.00 & 22.94 & 32.56 & 0.00 & 21.24\\
 & ResTune & {81.58} & 0.00 & \underline{55.80} & {69.75} & 0.00 & \underline{46.80} & {92.24} & 0.00 & \underline{55.70} & {84.62} & 0.00 & \underline{59.23} & {82.05} & 0.00 & \underline{54.38}\\
 & NCL & 46.64 & 0.00 & 31.90 & 27.12 & 0.00 & 18.20 & 69.65 & 0.00 & 41.30 & 4.60 & 0.00 & 3.51& 37.00 & 0.00 & 23.73\\
 & DTC & {68.57} & 0.00 & 46.90 & {61.10} & 0.00 & 41.00 & 31.20 & 0.00 & 18.50 & 81.36 & 0.00 & 56.93 & 60.56 & 0.00 & 40.83\\
  & GEM & 0.00 & {89.87} & 28.40 & 0.00 & {89.67} & 29.50 & 0.00 & {100.00} & 40.70 & 0.00 & {51.26} & 12.18 & 0.00 & {82.70} & 27.70 \\
 & ER-GNN & 0.00 & {89.87} & 28.40 & 0.00 & 85.41 & 28.10 & 0.00 & {100.00} & 40.70 & 0.00 & 46.70 & 11.10 & 0.00 & 80.50 & 27.08 \\
 & TWP & 0.00 & {90.82} & 28.70 & 0.00 & {89.06} & 29.30 & 0.00 & {100.00} & 40.70 & 0.00 & {51.86} & 12.33 & 0.00 & {82.94} & 27.76 \\
 & CPCA & {82.46} & 0.00 & \textbf{56.40} & {68.85} & 0.00 & {46.20} & {92.92} & 0.00 & \underline{55.10} & 30.05 & 0.00 & 22.91 & {68.57} & 0.00 & {45.15} \\
  & \cellcolor{lightgray}\textbf{SWORD} & \cellcolor{lightgray}60.67 & \cellcolor{lightgray}37.97 & \cellcolor{lightgray}{53.50} & \cellcolor{lightgray}58.87 & \cellcolor{lightgray}35.56 & \cellcolor{lightgray}\textbf{51.20} & \cellcolor{lightgray}{80.61} & \cellcolor{lightgray}{42.75} & \cellcolor{lightgray}\textbf{65.20} & \cellcolor{lightgray}{83.94} & \cellcolor{lightgray}39.02 & \cellcolor{lightgray}\textbf{63.99} & \cellcolor{lightgray}{71.02} & \cellcolor{lightgray}38.83 & \cellcolor{lightgray}\textbf{58.47}\\ \midrule

\multirow{8}{*}{GAT} & AutoNovel & 21.05 & 0.00 & 14.40 & 34.43 & 0.00 & 23.10 & {69.65} & 0.00 & \underline{41.30} & {83.27} & 0.00 & 28.29 & 52.10 & 0.00 & 26.77 \\
 & ResTune & {86.26} & 0.00 & 59.00 & {68.85} & 0.00 & 46.20 & {69.65} & 0.00 & \underline{41.30} & 7.41 & 0.00 & 5.65 & 58.04 & 0.00 & 38.04 \\
 & NCL & 19.01 & 0.00 & 13.00 & 11.48 & 0.00 & 7.70 & 30.35 & 0.00 & 18.00 & 8.83 & 0.00 & 6.73 & 17.42 & 0.00 & 11.36 \\
 & DTC & {70.18} & 0.00 & {48.00} & {68.26} & 0.00 & {45.80} & {69.65} & 0.00 & \underline{41.30} & {83.50} & 0.00 & \underline{58.45} & {72.90} & 0.00 & \underline{48.39} \\
  & GEM & 0.00 & 62.03 & 19.60 & 0.00 & 69.60 & 22.90 & 0.00 & {100.00} & 40.70 & 0.00 & {31.09} & 7.39 & 0.00 & 65.68 & 22.65 \\
 & ER-GNN & 0.00 & {75.32} & 23.80 & 0.00 & {74.47} & 24.50 & 0.00 & {100.00} & 40.70 & 0.00 & {31.09} & 7.39 & 0.00 & {70.22} & 24.10 \\
 & TWP & 0.00 & {82.59} & 26.10 & 0.00 & {86.93} & 28.60 & 0.00 & {100.00} & 40.70 & 0.00 & {31.09} & 7.39 & 0.00 & {75.15} & 25.70 \\
 & CPCA & {87.57} & 0.00 & \underline{59.90} & {72.43} & 0.00 & \underline{48.60} & {69.65} & 0.00 & 41.30 & 7.49 & 0.00 & 5.71 & 59.29 & 0.00 & 38.88 \\
  & \cellcolor{lightgray}\textbf{SWORD} & \cellcolor{lightgray}{74.71} & \cellcolor{lightgray}30.54 & \cellcolor{lightgray}\textbf{60.70} & \cellcolor{lightgray}{64.23} & \cellcolor{lightgray}44.79 & \cellcolor{lightgray}\textbf{56.30} & \cellcolor{lightgray}{79.79} & \cellcolor{lightgray}{45.70} & \cellcolor{lightgray}\textbf{65.50} & \cellcolor{lightgray}81.02 & \cellcolor{lightgray}{39.13} & \cellcolor{lightgray}\textbf{62.48} & \cellcolor{lightgray}{74.94} & \cellcolor{lightgray}40.04 & \cellcolor{lightgray}\textbf{61.25} \\ \midrule

\multirow{8}{*}{GraphSAGE} & AutoNovel & 24.27 & 0.00 & 16.60 & 12.52 & 0.00 & 8.40 & \underline{69.81} & 0.00 & 41.40 & 68.41 & 0.00 & 52.15 & 43.75 & 0.00 & 29.64 \\
 & ResTune & {80.48} & 0.00 & 53.60 & {70.38} & 0.00 & \underline{47.20} & 69.65 & 0.00 & 41.30 & 26.12 & 0.00 & 19.91 & 61.66 & 0.00 & 40.50 \\
 & NCL & 19.01 & 0.00 & 13.00 & 26.97 & 0.00 & 18.10 & 30.35 & 0.00 & 18.00 & 3.29 & 0.00 & 2.51 & 19.91 & 0.00 & 12.90 \\
 & DTC & {76.32} & 0.00 & {52.20} & {52.46} & 0.00 & {35.20} & {85.33} & 0.00 & {50.60} & {82.22} & 0.00 & \underline{62.68} & {74.08} & 0.00 & \underline{50.17} \\
 & GEM & 0.00 & {90.19} & 28.50 & 0.00 & {89.97} & 29.60 & 0.00 & {100.00} & 40.70 & 0.00 & {51.26} & 12.18 & 0.00 & {82.86} & 27.75 \\
 & ER-GNN & 0.00 & 86.08 & 27.20 & 0.00 & 88.15 & 29.00 & 0.00 & {100.00} & 40.70 & 0.00 & {51.38} & 12.21 & 0.00 & 81.40 & 27.28 \\
 & TWP & 0.00 & {88.29} & 27.90 & 0.00 & {89.67} & 29.50 & 0.00 & {100.00} & 40.70 & 0.00 & {51.38} & 12.21 & 0.00 & {82.34} & 27.58 \\
 & CPCA & {80.12} & 0.00 & \underline{54.80} & {69.45} & 0.00 & \underline{46.60} & 91.23 & 0.00 & \underline{54.10} & 7.49 & 0.00 & 5.71 & 62.07 & 0.00 & 40.30 \\
 & \cellcolor{lightgray}\textbf{SWORD}  & \cellcolor{lightgray}{69.07} & \cellcolor{lightgray}36.71 & \cellcolor{lightgray}\textbf{57.50} & \cellcolor{lightgray}{59.61} & \cellcolor{lightgray}26.75 & \cellcolor{lightgray}\textbf{48.80} & \cellcolor{lightgray}64.92 & \cellcolor{lightgray}{66.83} & \cellcolor{lightgray}\textbf{64.10} & \cellcolor{lightgray}{84.80} & \cellcolor{lightgray}45.65 & \cellcolor{lightgray}\textbf{64.39} & \cellcolor{lightgray}{69.60} & \cellcolor{lightgray}43.99 & \cellcolor{lightgray}\textbf{58.70} \\ \bottomrule
\end{tabular}}
\end{table*}

\subsection{Comparison with SOTA Methods}
Under our practical setting for node classification (i.e., NC-NCD), we compare SWORD with the NCD baseline methods and record the performance on the old, new and all categories after unified classification using the joint classifier $h^a$ on all unlabeled samples, as shown in Table \ref{table:comp}.
The best results are highlighted in bold, while the runner-ups are underlined.
From the comprehensive views, we make the following observations and analyses.

\vspace{-1.5mm}
\subsubsection{Analysis of Performance Comparison}
First, our SWORD stands out among the baseline methods as it effectively balances the classification of old and new categories.
After learning old category nodes on task $\mathcal{T}^l$ and learning new category nodes on task $\mathcal{T}^u$, SWORD successfully distinguishes between old and new categories using the joint classifier and performs
better on the classification of all category nodes.

In contrast, most of the NCD baseline methods struggle to differentiate new nodes using task-agnostic joint classifiers, prioritizing the maintenance of performance on old classes. Although these NCD methods fail to classify new category nodes in the NC-NCD setting, they still achieve better overall performance by relying on the classification of old categories.
We acknowledge that the performance on all categories is largely influenced by the classification accuracy on the old categories.
Further analysis with confusion matrices of this phenomenon will be conducted in Appendix~\ref{app:matrix}.

Regarding the aforementioned graph incremental learning methods, they require labeled nodes for both new and old categories during phase 2 when learning new categories. 
Therefore, CPCA in the class-IL setting fails to differentiate new categories in the absence of new category labels.
In the context of task-IL methods, task-ids need to be specified for evaluation under the assumption of known category membership (whether a category belongs to the new or old set).
By leveraging the powerful node representation learning capabilities of GNNs, these methods excel in the classification of new category nodes.
However, in the NC-NCD setting where labeled data for old categories is absent, they struggle to differentiate among old categories, resulting in weak performance across all categories.
Due to the loss of old category node classification capability and the inability to balance the classification performance between new and old categories, 
the aforementioned task-IL methods prove inadequate for NC-NCD tasks.

In addition, we utilize average accuracy (AA) and average forgetting (AF) \cite{zhang2022cglb} to assess the average performance in learning new classes and the resistance to forgetting in Appendix~\ref{app:aaaf}.

\vspace{-1.5mm}
\subsubsection{Performance on Multiple Backbones}
We implement SWORD with multiple backbones and evaluate the node classification accuracy for both old and new categories.
Our results show that SWORD consistently outperforms other SOTA methods in achieving a balanced performance on old and new categories. Specifically, GraphSAGE and GCN exhibit similar performance on multiple datasets.
GAT, with its attention mechanism, stands out in accurately classifying both old and new category nodes.

We also visualize the Cora dataset's original graph and the embedding of graphs learned by SWORD and other baseline methods via t-SNE in Appendix~\ref{app:tsne}.

\subsection{Ablation Studies}

\begin{table}[!t]
\caption{Comparison on node classification accuracy with task-specific head. Old: joint head for old classes; New-N: novel head for new classes.}
\label{table:head-novel}
\centering
\resizebox{0.9\linewidth}{!}{
\begin{tabular}{@{}l|ccc|ccc@{}}
\toprule
\multirow{2}{*}{Methods} & \multicolumn{3}{c|}{Cora} & \multicolumn{3}{c}{Citeseer} \\ \cmidrule(l){2-7}
 & Old & New-N & \textbf{All} & Old & New-N & \textbf{All} \\ \midrule

AutoNovel & 19.01 & 21.84 & 13.00 & 11.48 & {48.63} & 7.70 \\
NCL & 46.64 &{47.15} & 31.90 & 27.12 & {48.63} & 18.20 \\
DTC & {68.57} & 19.94 & \underline{46.90} & {61.10} & {45.90} & \underline{41.00} \\
\textbf{SWORD} & {60.67} & {48.42} & \textbf{53.50} & {58.87} & 45.59 & \textbf{51.20} \\ \bottomrule
\end{tabular}}
\vspace{-0.2cm}
\end{table}

In this section, we delve deeper into the effectiveness of the baselines on NCD tasks and analyze the effectiveness of different components within our framework.

\vspace{-1.5mm}
\subsubsection{Effectiveness of the Baselines on NCD tasks}
As shown in Table \ref{table:head-novel}, the baseline methods and our SWORD achieve commendable classification accuracy for both old and new category nodes, primarily attributed to the utilization of task-specific novel classification head for new classes. This underscores the efficacy of these SOTA methods in addressing NCD tasks. It is important to emphasize that their effectiveness is contingent upon the specified classification heads, a distinction from our NC-NCD setting.

Besides, the results in Table \ref{table:comp} reveals that the baseline methods with joint head struggle to effectively classify new categories.
Nevertheless, in the more realistic NC-NCD setting, the previous NCD methods might lose their capacity on NCD tasks for failing to accurately discern whether a node pertains to old or new categories. In contrast, this limitation is addressed by our SWORD.

In addition, five different loss functions are proposed by SWORD, and ablation studies of the effectiveness of $\mathcal{L}_{Pseudo}$,  $\mathcal{L}_{Perturb}$, $\mathcal{L}_{Self}$, $\mathcal{L}_{Replay}$ and $\mathcal{L}_{Distill}$ (abbreviated as $\mathcal{L}_{Ps}$, $\mathcal{L}_{Pt}$, $\mathcal{L}_{Se}$, $\mathcal{L}_{Re}$, and $\mathcal{L}_{Ds}$ respectively) to achieve the NC-NCD task is shown in Table \ref{table:abla}.
In general, the best classification results of the model, i.e., the ability to maintain the performance of old classes while learning the new classes, are obtained when the five loss functions are combined as new class nodes emerge.
Different loss functions optimize the classification results in different ways, and removing any loss function can have a significant impact on the performance.

\vspace{-1.5mm}
\subsubsection{Effectiveness of Prototype Replay and Distillation}
Table \ref{table:abla} shows that the removal of either $\mathcal{L}_{Replay}$ or $\mathcal{L}_{Distill}$ causes SWORD to completely forget the old categories. This outcome aligns with our expectations, as the feature extractor deviates from its original configuration after training on the new task $\mathcal{T}^u$, thus leading the extracted node representations of the old classes to diverge from their prototypes.

Furthermore, we observed that SWORD, when subjected to the ablation of prototype replay and distillation, exhibits superior performance in classifying new categories on the  Citeseer dataset.
This enhanced performance can be attributed to the relatively smaller number of novel node categories present in the dataset. Following the completion of task $\mathcal{T}^u$, the model with the ablated prototype replay and distillation mechanisms tends to classify all categories as new categories.

\vspace{-1.5mm}
\subsubsection{Effectiveness of Self-training}
As shown in the 4th row of Table \ref{table:abla}, SWORD, when lacking self-training, faces challenges in learning new category nodes, while the joint classifier maintains its performance on old category nodes.
Although the model exhibits excellent classification performance on both old and all categories at this stage, it tends to classify all categories as old categories. In other words, the overall higher performance of the model on all categories relies heavily on the accurate classification of the old categories. This observation is not consistent with the practical NC-NCD setting, where it is essential to achieve a balanced performance between old and new categories. Thus, the self-training strategy employed by SWORD proves effective in addressing this issue.

A similar phenomenon can be observed with the confusion matrix in Figure \ref{fig:confusion-matrix} in Appendix~\ref{app:matrix},
further emphasizing the complexity of balancing the performance on old and new categories in the NC-NCD setting.
Additionally, when removing $\mathcal{L}_{Replay}$ and $\mathcal{L}_{Distill}$ along with $\mathcal{L}_{Self}$, the model not only fails to learn the new categories but also further degrades its performance on the old categories.
This aligns with our expectations and demonstrates the effectiveness of $\mathcal{L}_{Replay}$ and $\mathcal{L}_{Distill}$ in preventing the forgetting of old categories.

\begin{table}[!t]
\vspace{-0.2cm}
\centering
\caption{Ablation study of loss functions.}
\label{table:abla}
\resizebox{\linewidth}{!}{
\begin{tabular}{ccccc|ccc|ccc}
\toprule
\multicolumn{5}{c|}{Loss functions} & \multicolumn{3}{c|}{Cora} & \multicolumn{3}{c}{Citeseer} \\
\midrule
$\mathcal{L}_{Ps}$ & $\mathcal{L}_{Pt}$ & $\mathcal{L}_{Se}$ & $\mathcal{L}_{Re}$ & $\mathcal{L}_{Ds}$ & Old(4) & New(3) & All(7) & Old(4) & New(2) & All(6)\\
\midrule
$\checkmark$ & $\checkmark$ & $\checkmark$ & $\checkmark$ &  & 0.00 & 32.59 & 10.30 & 0.00 & \textbf{51.37} & 16.90 \\
$\checkmark$ & $\checkmark$ & $\checkmark$ &  & $\checkmark$ & 0.00 & 37.76 & 11.30 & 1.49 & \underline{44.38} & 15.60 \\
$\checkmark$ & $\checkmark$ & $\checkmark$ &  &  & 0.00 & \textbf{51.90} & 16.40 & 0.00 & \textbf{51.37} & 16.90\\
$\checkmark$ & $\checkmark$ &  & $\checkmark$ & $\checkmark$ & \textbf{82.16} & 0.00 & \textbf{56.20} & \textbf{69.60} & 0.00 & {46.70}  \\
$\checkmark$ & $\checkmark$ &  &  &  & 35.53 & 0.00 & 24.30 & 34.43 & 0.00 & 23.10  \\
$\checkmark$ & & $\checkmark$ & $\checkmark$ & $\checkmark$ & {63.74} & 27.22 & 52.20 & \underline{59.02} & 32.83 & \underline{50.40} \\
 & $\checkmark$ & $\checkmark$ & $\checkmark$ & $\checkmark$ & \underline{72.07} & 8.98 & 51.00 & 58.16 & 26.99 & 48.00 \\
\midrule
$\checkmark$ & $\checkmark$ & $\checkmark$ & $\checkmark$ & $\checkmark$ & {60.67} & \underline{37.97} & \underline{53.50} & {58.87} & 35.56 & \textbf{51.20}\\
\bottomrule
\end{tabular}}
\vspace{-0.2cm}
\end{table}

\subsection{Parameter Analysis}
In this section, we present extensive experiments to analyze the sensitivity of our SWORD to the ramp-up function weights $\beta_1$ and $\beta_2$, noise perturbation $\eta$ and hyper-parameter $\lambda$.
The results are shown in Figure \ref{fig:sensi}.
In the Appendix\ref{app:layers}, we conduct further experiments to analyze the impact of the number of layers in the GNN backbones on classification performance.

\vspace{-1.5mm}
\subsubsection{Effect of the Weights of Ramp-up Functions}
The coefficients $\beta_1$ and $\beta_2$ are the weights adjusted by the ramp-up function, represented as $\beta_1 = \alpha_1 \cdot \mathrm{rampup}({epoch})$,
where $\alpha_1$ can be flexibly adjusted to the loss function at different training stages. $\beta_2$ is in the same vein.
Therefore, we tuned adjustable coefficients $\alpha_1$ and $\alpha_2$ to examine their impact on the model's performance in the NC-NCD task.
Figure \ref{fig:sensi}(a) demonstrates that as $\alpha_1$ increases, the performance of SWORD gradually improves on the new categories, while the effect on the old and all categories diminishes.
Conversely, $\alpha_2$ has a minor influence on the model's performance. Increasing the noise perturbation by raising $\alpha_2$ results in a slight improvement in the learning of new category nodes, while it has minimal effect on old categories.

\vspace{-1.5mm}
\subsubsection{Effect of Gaussian Noise Perturbation}
The coefficient $\eta$ is used to measure the extent to which random Gaussian noise is added to the new class data in the joint training head. As shown in Figure \ref{fig:sensi}(c), a small increase in the perturbation of Gaussian noise, for example, $\eta$ ranging from 0.05 to 0.2, can improve the performance on the new classes while maintaining the performance of the old classes. In contrast, larger values of $\eta$ lead to higher fluctuations in the classification performance of the new categories.
This observation aligns with our intuition.

\vspace{-1.5mm}
\subsubsection{Effect of Loss Balance Factor}
\begin{figure}[!t]
\vspace{-0.2cm}
\centering
\subfloat[Analysis of $\alpha_1$]{
\includegraphics[width=.22\textwidth]{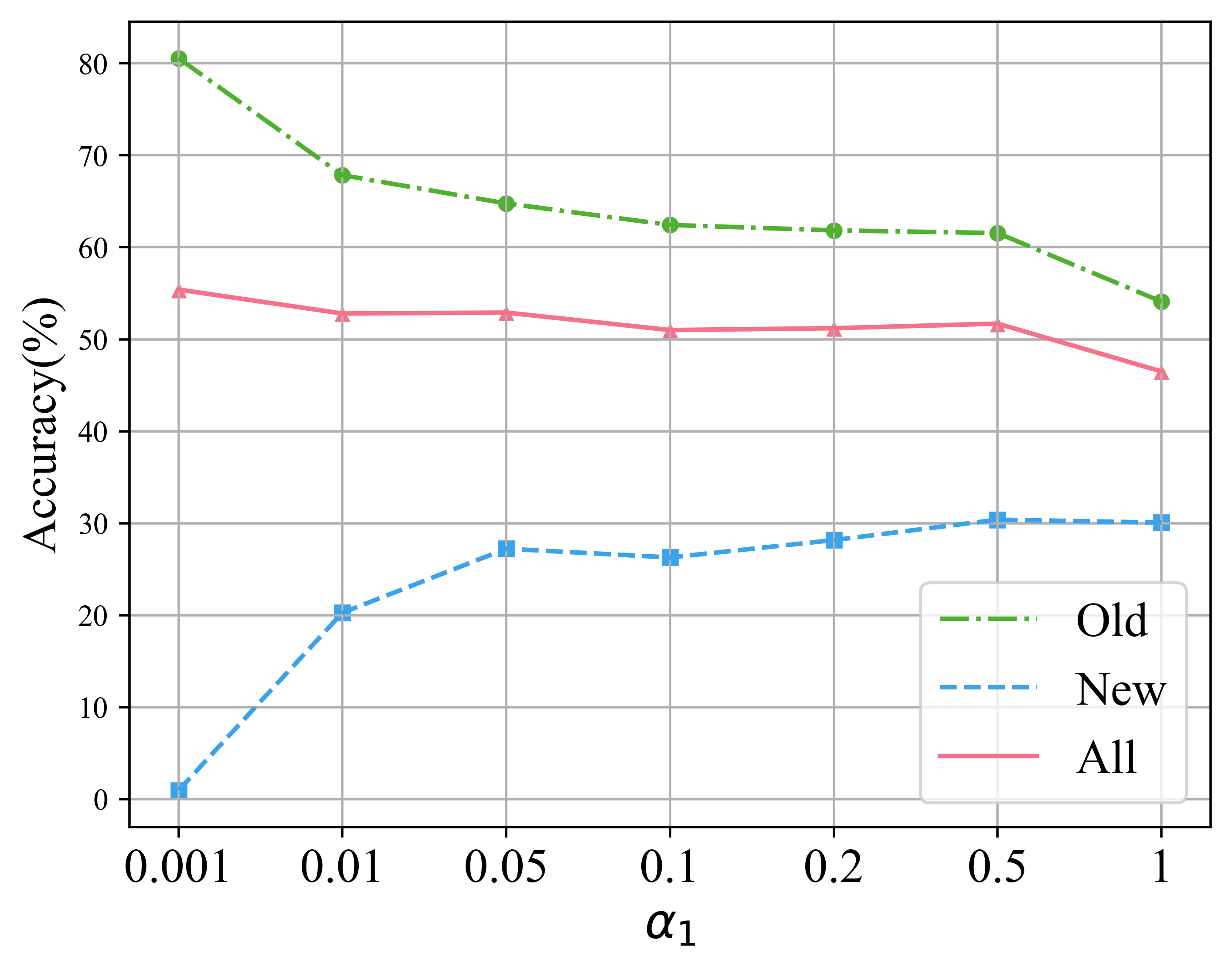}}
\subfloat[Analysis of $\alpha_2$]{
\includegraphics[width=.22\textwidth]{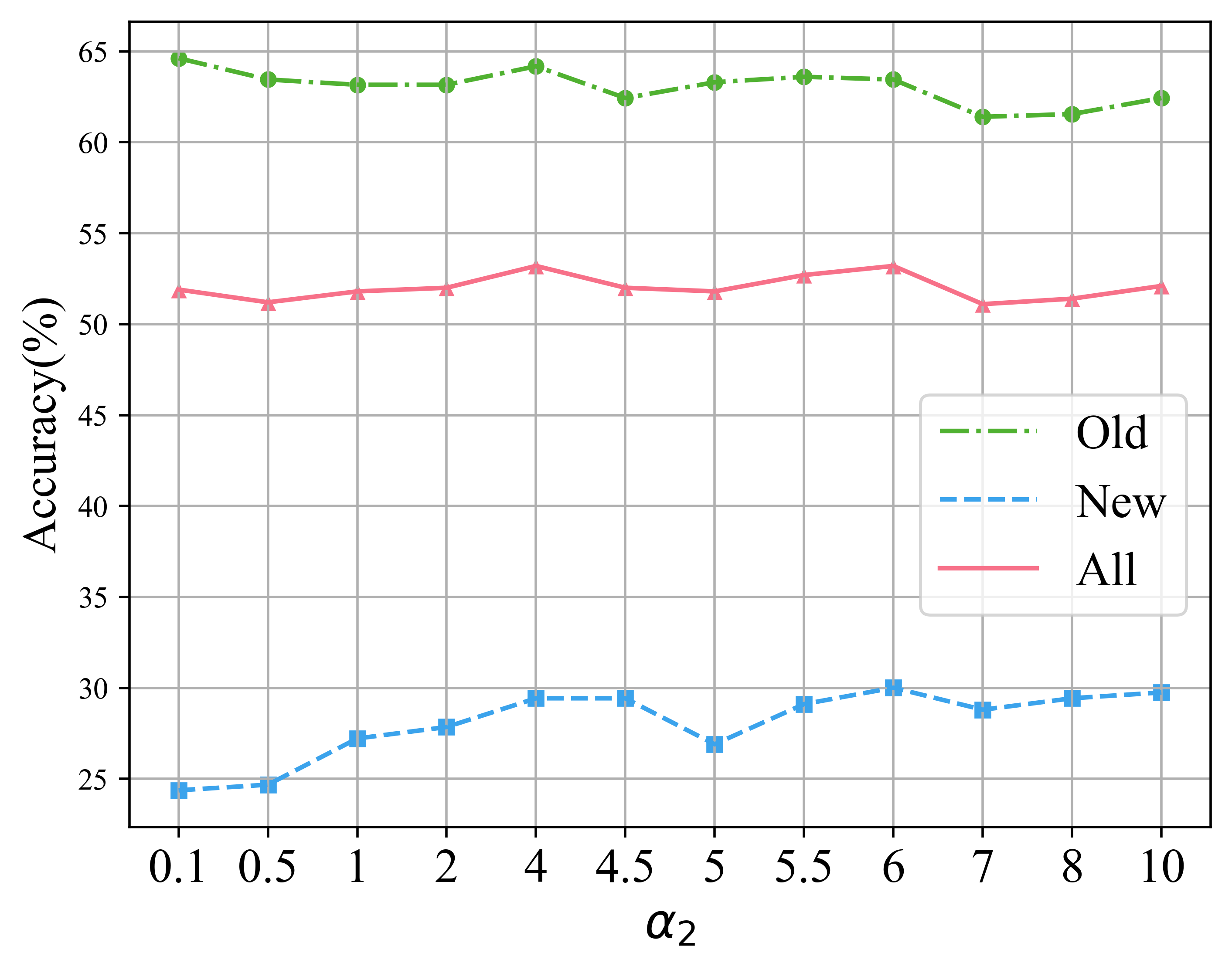}} \\
\subfloat[Analysis of $\eta$]{
\includegraphics[width=.22\textwidth]{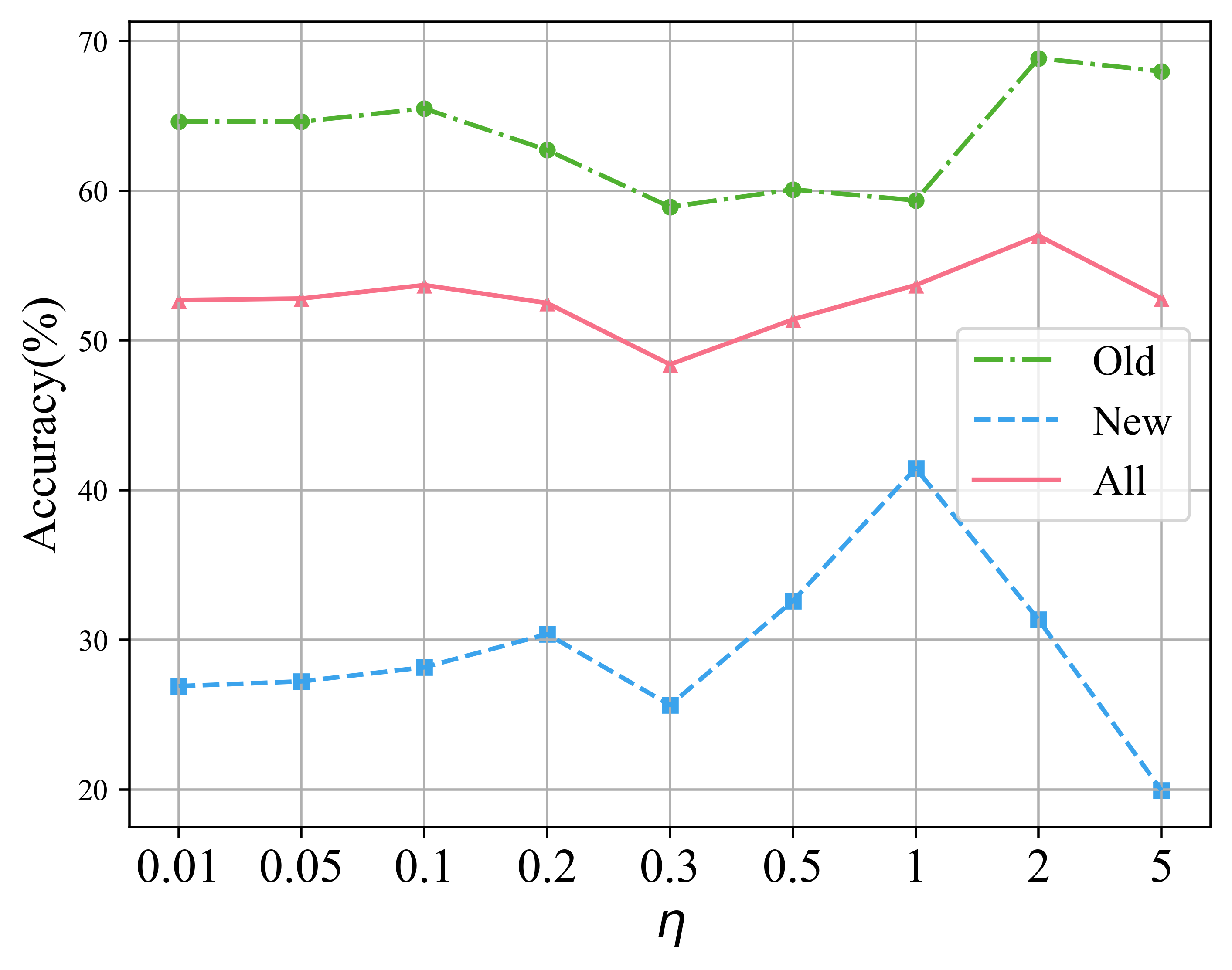}}
\subfloat[Analysis of $\lambda$]{
\includegraphics[width=.22\textwidth]{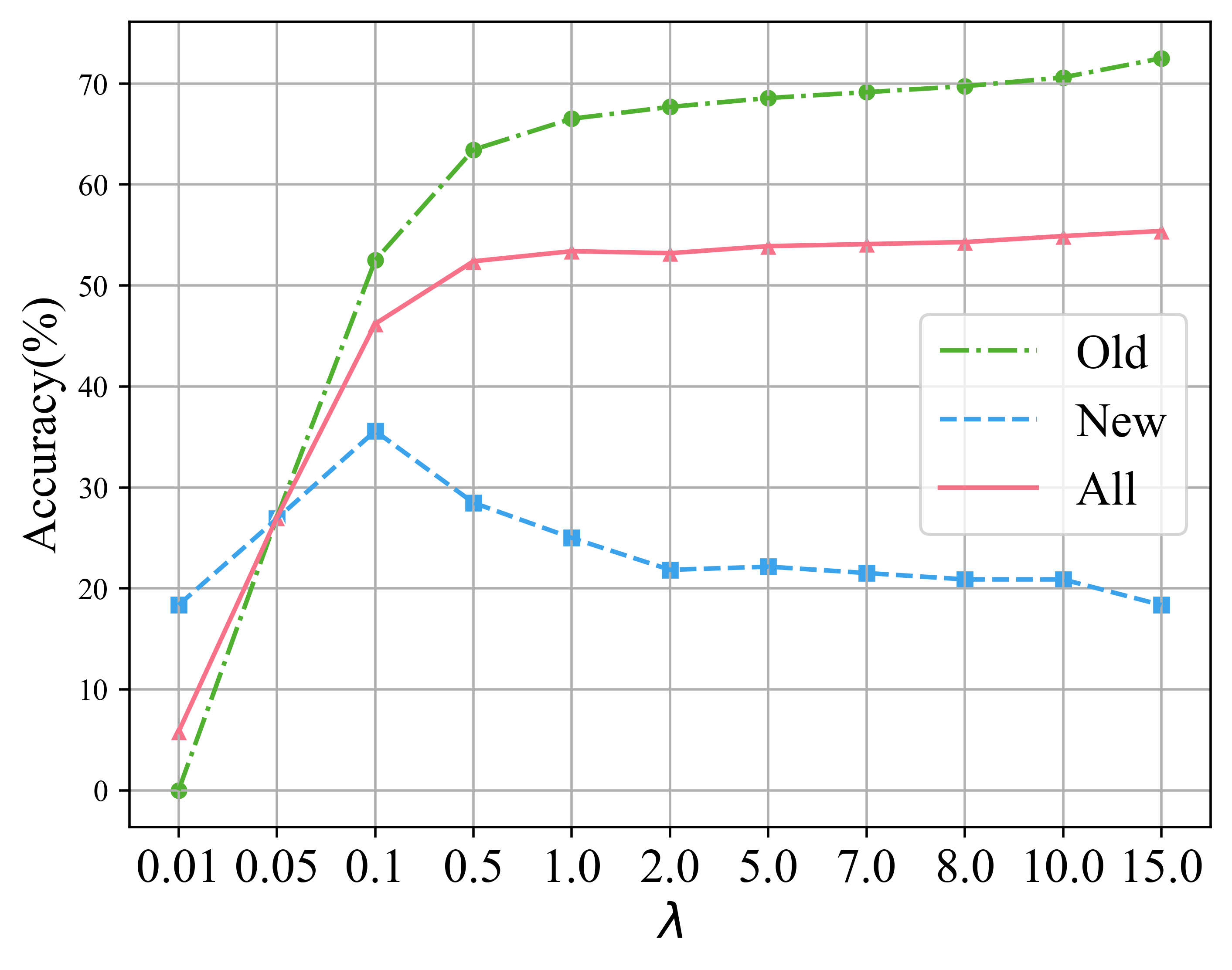}}
\vspace{-0.2cm}
    \caption{Sensitivity analysis w.r.t. parameter $\alpha_1$, $\alpha_2$, $\eta$ and $\lambda$.}
  \label{fig:sensi}
\end{figure}

The hyper-parameter $\lambda$ is an essential loss balance factor to weigh the learning ability of the model at the old and new category nodes, preventing SWORD from completely failing to learn the new categories or completely forgetting the old categories. As we found in Figure \ref{fig:sensi}(d), when $\lambda$ is small, the model's forgetting of the old categories is catastrophic, while when $\lambda$ is large enough the performance improvement on the old categories slows down and is replaced by a performance decrease on the new categories. This demonstrates the important role of $\lambda$ in balancing the performance of the old and new categories.

\section{Conclusion}
In this work, we propose a novel NC-NCD setting that distinguishes itself from previous NCD tasks.
In real-world scenarios, existing NCD methods fail to adequately balance the learning performance between the old and new categories in a task-agnostic manner.
To address this limitation, we introduce a new framework called SWORD to enhance the node classification task in the NC-NCD setting.
SWORD employs a self-training strategy to train task-agnostic joint classifiers capable of classifying all previously seen categories, eliminating the need for explicit task indicators.
Additionally, feature-level prototype replay as well as distillation are utilized to prevent forgetting.
Comprehensive experiments demonstrate the potential and adaptability of the SWORD method for NC-NCD tasks compared to other SOTA methods. 
A practical future endeavor is to embark on developing a framework that does not necessitate prior knowledge of the number of novel classes and to extend the NC-NCD framework to multiple stages NCD. We will also discuss this within the generalized category discovery setting, where both unlabeled old and new category data are available which reflects real-world scenarios.

\begin{acks}
    This research was supported by NSFC (Grant No. 61932002).
\end{acks}

\bibliographystyle{ACM-Reference-Format}
\bibliography{nc2d.bib}

\clearpage
\appendix
\section{APPENDIX}

\subsection{Overall Process of SWORD} \label{app:algo}

The overall process of SWORD is shown in Algorithm \ref{algo:overall} as follows.

\begin{algorithm}[!t]
  \caption{The overall process of SWORD in NC-NCD setting}
  \label{algo:overall}
  \SetAlgoLined
  \KwIn{Graph dataset $\mathcal{G}$, Number of NCD-training epochs $E$, coefficients $\beta_1$, $\beta_2$, $\lambda$.}
  \KwOut{Trained model $f=h^a \circ g$ on task $\mathcal{T}^u$}
  \tcp{Data split}
  Split dataset $\mathcal{G}$ into $\mathcal{G}^l$ and $\mathcal{G}^u$.\\
  Parameter initialization\;
  \tcp{Pre-training}
  Pre-train the model on $\mathcal{G}^l$ to calculate $\mathcal{L}_{ce}$.\\
  \tcp{NCD training (SWORD)}
  \For{$e = 1, \ldots, E$}{
Calculate the pairwise similarity $s_{ij} \leftarrow$ Eq.(\ref{equa:pair_simi}) and pairwise pseudo labels $\tilde{y}_{ij}^u \leftarrow$\ Eq.(\ref{equa:pair_pseudo})\;
Calculate the losses $\mathcal{L}_{Pseudo} \leftarrow$ Eq.(\ref{equa:loss_pseudo}), $\mathcal{L}_{Self} \leftarrow$ Eq.(\ref{equa:loss_self}) and $\mathcal{L}_{Perturb} \leftarrow$ Eq.(\ref{equa:loss_perturb})\;
Obtain prototype $\boldsymbol{\mu_c}$ and the variance $\boldsymbol{\nu_c^2} \leftarrow$ Eq.(\ref{equa:mu_nu}) on task $\mathcal{T}^l$\;
Calculate the losses $\mathcal{L}_{Replay} \leftarrow$ Eq.(\ref{equa:loss_replay}) and $\mathcal{L}_{Distill} \leftarrow$ Eq.(\ref{equa:loss_kd})\;
Update model parameters to minimize $\mathcal{L} \leftarrow$ Eq.(\ref{equa:loss_overall}).\\
  }
  \tcp{Inference and Evaluation}
  Evaluates the performance of all classes in $\mathcal{G}_{test}$ during the inference phase.\\
\end{algorithm}

\subsection{Datasets} \label{app:datasets}
In this section, we describe more details on the datasets used for this paper. 

\begin{itemize}
\item Cora, Citeseer and Pubmed are citation networks following the standard train/validation/test split in line with \cite{kipf2016semi}.

\item Wiki-CS consists of nodes corresponding to Computer Science articles, with edges based on hyperlinks and 10 classes representing different branches of the field.
We split this dataset according to \cite{mernyei2020wiki}.
\end{itemize}

In the novel NC-NCD, the pre-training phase employs labeled node features exclusively from the old categories, whereas the NCD-training phase involves unlabeled node features specific to the new categories.
Therefore, the splitted datasets
are further randomly divided into old and new classes according to node labels, with node-ids recorded to facilitate the reproduction and comparison of subsequent methods.
To elaborate, distinct categories of node features are masked during different training phases driven by node-ids, thereby effectuating the division of graph data nodes for diverse tasks.

\begin{table}[!t]
\caption{Dataset statistical description and split.}
\label{table:dataset}
\resizebox{0.9\linewidth}{!}{
\begin{tabular}{@{}lccccc@{}}
\toprule
\multirow{2}{*}{Dataset} & \multirow{2}{*}{Nodes} & \multirow{2}{*}{Edges} & \multirow{2}{*}{Features} & \multicolumn{2}{c}{Class split} \\ \cmidrule(lr){5-6}
 &  &  &  & Old classes& New classes \\ \midrule
Cora & 2,708 & 10,556 & 1,433 & 4 & 3 \\
Citeseer & 3,327 & 9,104 & 3,703 & 4 & 2 \\
Pubmed & 19,717 & 88,648 & 500 & 2 & 1 \\
Wiki-CS & 11,707 & 216,123 & 300 & 7 & 3 \\ \bottomrule
\end{tabular}
}
\end{table}

\subsection{Baselines} \label{app:baselines}
Here are descriptions and implementation details of the baseline methods used in the experiments:

\textbf{NCD Methods:}
\begin{itemize}
    \item \textbf{AutoNovel} \cite{han2020automatically} initiates a self-supervised pre-training phase using both labeled and unlabeled data to initialize the network, followed by supervised training exclusively with labeled data.
    \item \textbf{ResTune} \cite{liu2022residual} partitions data features into two components: basic features, aimed at retaining information from labeled data, and residual features, dedicated to adjusting information derived from unlabeled data. The optimization of residual features is achieved through clustering objectives.
    \item \textbf{NCL} \cite{zhong2021neighborhood} exploits local neighborhoods within the embedding space, empowering the model to assimilate knowledge from a broader set of positive samples, thereby elevating clustering accuracy.
    \item \textbf{DTC} \cite{han2019learning} employs the clustering algorithm to identify k-means centers for unlabeled data and subsequently estimates a potential target distribution to refine these centers.
\end{itemize}

\textbf{task-IL Methods:}
\begin{itemize}
\item \textbf{GEM} \cite{lopez2017gradient} stores representative data in episodic memory. During the learning process, GEM modifies the gradients of the current task by incorporating the gradient calculated with the stored data, preventing an increase in the loss of previous tasks.
\item \textbf{ER-GNN} \cite{zhou2021overcoming} represents a memory replay-based graph incremental learning method integrating memory replay into GNNs by preserving representative nodes selected from previous tasks.
\item \textbf{TWP} \cite{liu2021overcoming} is a regularization-based graph incremental learning method that introduces a penalty to preserve the topological information of previous graphs.
\end{itemize}

\textbf{class-IL Methods:}
\begin{itemize}
\item \textbf{CPCA} \cite{ren2023incremental} is a class-IL method on graphs, that employs class prototype augmentation, creating virtual classes through the combination of current prototypes to facilitate the learning of new category nodes.
\end{itemize}

\begin{table*}[!t]
\caption{Comparison of AA and AF results with different SOTA methods and GNN backbones in NC-NCD setting. The symbol $\uparrow$ indicates higher is better.}
\label{table:aaaf}
\centering
\resizebox{0.7\linewidth}{!}{
\begin{tabular}{@{}c|c|cc|cc|cc|cc|cc@{}}
\toprule

 \multirow{2}{*}{Backbone} & \multirow{2}{*}{Methods} & \multicolumn{2}{c|}{Cora} & \multicolumn{2}{c|}{Citeseer} & \multicolumn{2}{c|}{Pubmed} & \multicolumn{2}{c|}{Wiki-CS} & \multicolumn{2}{c}{Average} \\ \cmidrule(l){3-12} 
&  & $\uparrow$ AA &{$\uparrow$ AF} & $\uparrow$ AA & {$\uparrow$ AF} & $\uparrow$ AA & {$\uparrow$ AF} & $\uparrow$ AA & {$\uparrow$ AF} & $\uparrow$ AA & $\uparrow$ AF \\\midrule

\multirow{8}{*}{GCN} & \multicolumn{1}{c|}{AutoNovel} & 9.51 & -66.37 & 5.74 & -58.27 & 34.83 & -22.93 & 15.05 & -53.67 & 16.28 & -50.31 \\
 & \multicolumn{1}{c|}{ResTune} & 40.79 & \underline{-3.80} & 34.88 & \textbf{0.00} & 46.12 & \textbf{-0.34} & \underline{42.31} & \textbf{2.02} & 41.02 & \textbf{-0.53} \\
 & \multicolumn{1}{c|}{NCL} & 23.32 & -38.74 & 13.56 & -42.63 & 34.83 & -22.93 & 2.30 & -78.00 & 18.50 & -45.58 \\
 & \multicolumn{1}{c|}{DTC} & 34.29 & {-16.81} & 30.55 & {-8.65} & 15.60 & -61.38 & 40.68 & -1.24 & 30.28 & -22.02 \\
 & \multicolumn{1}{c|}{GEM} & 44.94 & -85.38 & \underline{44.84} & -69.75 & \underline{50.00} & -92.58 & 25.63 & -82.60 & {41.35} & -82.58 \\
 & \multicolumn{1}{c|}{ER-GNN} & 44.94 & -85.38 & 42.71 & -69.75 & \underline{50.00} & -92.58 & 23.35 & -82.60 & 40.25 & -82.58 \\
 & \multicolumn{1}{c|}{TWP} & \underline{45.41} & -85.38 & 44.53 & -69.75 & \underline{50.00} & -92.58 & 25.93 & -82.60 & \underline{41.47} & -82.58 \\
 & CPCA & 41.23 & \textbf{-2.92} & 34.43 & \underline{-0.90} & 46.46 & \textbf{0.34} & 15.03 & -52.55 & 34.29 & {-14.01} \\
  & \cellcolor{lightgray}{\textbf{SWORD}} & \cellcolor{lightgray}\textbf{49.32} & \cellcolor{lightgray}-24.71 & \cellcolor{lightgray}\textbf{47.22} & \cellcolor{lightgray}-10.88 & \cellcolor{lightgray}\textbf{61.68} & \cellcolor{lightgray}\underline{-11.97} & \cellcolor{lightgray}\textbf{61.48} & \cellcolor{lightgray}\underline{1.34} & \cellcolor{lightgray}\textbf{54.92} & \cellcolor{lightgray}\underline{-11.56} \\ \midrule

\multirow{8}{*}{GAT} & \multicolumn{1}{c|}{AutoNovel} & 10.53 & -64.18 & 17.22 & -26.23 & 34.83 & \underline{-21.41} & 41.64 & -0.23 & 26.05 & -28.01 \\
 & \multicolumn{1}{c|}{ResTune} & 43.13 & \underline{1.03} & 34.43 & \underline{8.91} & 34.83 & \underline{-21.41} & 3.71 & -76.09 & 29.02 & -22.07 \\
 & \multicolumn{1}{c|}{NCL} & 9.51 & -66.22 & 5.74 & -49.18 & 15.18 & -60.71 & 4.42 & -74.67 & 8.71 & -62.70 \\
 & \multicolumn{1}{c|}{DTC} & 35.09 & {-15.05} & 34.13 & {7.60} & 34.83 & \underline{-21.41} & \underline{41.75} & \textbf{0.00} & 36.45 & \underline{-7.22} \\
 & \multicolumn{1}{c|}{GEM} & 31.02 & -85.23 & 34.80 & -60.66 & \underline{50.00} & -91.06 & 15.55 & -83.50 & 32.84 & -80.11 \\
 & \multicolumn{1}{c|}{ER-GNN} & 37.66 & -85.23 & 37.24 & -60.66 & \underline{50.00} & -91.06 & 15.55 & -83.50 & 35.11 & -80.11 \\
 & \multicolumn{1}{c|}{TWP} & {41.30} & -85.23 & \underline{43.47} & -60.66 & \underline{50.00} & -91.06 & 15.55 & -83.50 & \underline{37.58} & -80.11 \\
  & CPCA & \underline{43.79} & \textbf{2.34} & 36.22 & \textbf{11.77} & 34.83 & \underline{-21.41} & 3.75 & -76.01 & 29.64 & -20.83 \\
  & \cellcolor{lightgray}{\textbf{SWORD}} & \cellcolor{lightgray}\textbf{52.63} & \cellcolor{lightgray}{-10.52} & \cellcolor{lightgray}\textbf{54.51} & \cellcolor{lightgray}{3.57} & \cellcolor{lightgray}\textbf{62.75} & \cellcolor{lightgray}\textbf{-11.27} & \cellcolor{lightgray}\textbf{60.08} & \cellcolor{lightgray}\underline{-2.48} & \cellcolor{lightgray}\textbf{57.49} & \cellcolor{lightgray}\textbf{-5.18} \\ \midrule

\multirow{8}{*}{GraphSAGE} & \multicolumn{1}{c|}{AutoNovel} & 12.14 & -61.40 & 6.26 & -56.78 & 34.91 & {-21.59} & 34.21 & -16.10 & 21.88 & -38.97 \\
 & \multicolumn{1}{c|}{ResTune} & 40.02 & \textbf{-5.19} & 35.19 & \textbf{1.08} & 34.83 & -21.75 & 13.06 & -58.39 & 30.83 & -21.06 \\
 & \multicolumn{1}{c|}{NCL} & 9.51 & -66.66 & 13.49 & -42.33 & 15.18 & -61.05 & 1.65 & -81.22 & 9.95 & -62.82 \\
 & \multicolumn{1}{c|}{DTC} & 38.16 & {-9.35} & 26.23 & {-16.84} & 42.67 & \underline{-6.07} & \underline{41.11} & \underline{-2.29} & 37.04 & \textbf{-8.64} \\
 & \multicolumn{1}{c|}{GEM} & \underline{45.10} & -85.67 & \textbf{44.99} & -69.30 & \underline{50.00} & -91.40 & 25.63 & -84.51 & \underline{41.43} & -82.72 \\
 & \multicolumn{1}{c|}{ER-GNN} & 43.04 & -85.67 & 44.08 & -69.30 & \underline{50.00} & -91.40 & 25.69 & -84.51 & 40.70 & -82.72 \\
 & \multicolumn{1}{c|}{TWP} & 44.15 & -85.67 & \underline{44.84} & -69.30 & \underline{50.00} & -91.40 & 25.69 & -84.51 & 41.17 & -82.72 \\ 
 & CPCA & 40.06 & \underline{-5.55} & 34.73 & \underline{0.15} & 45.62 & \textbf{-0.17} & 3.75 & -77.02 & 31.04 & -20.65 \\
  & \cellcolor{lightgray}\textbf{SWORD} & \cellcolor{lightgray}\textbf{52.89} & \cellcolor{lightgray}{-16.60} & \cellcolor{lightgray}43.18 & \cellcolor{lightgray}{-9.69} & \cellcolor{lightgray}\textbf{65.88} & \cellcolor{lightgray}-26.48 & \cellcolor{lightgray}\textbf{65.23} & \cellcolor{lightgray}\textbf{0.29} & \cellcolor{lightgray}\textbf{56.79} & \cellcolor{lightgray}\underline{-13.12} \\
\bottomrule
\end{tabular}}
\end{table*}

These previous NCD methods are originally implemented on image data. Therefore, we replicate them based on GNN backbones in our NC-NCD setting.
GEM \cite{lopez2017gradient} is a popular incremental learning method for Euclidean data, and is considered a baseline on graphs in many studies \cite{zhang2022cglb, zhou2021overcoming, liu2021overcoming}.
ER-GNN \cite{zhou2021overcoming} and TWP \cite{liu2021overcoming} are both classical graph incremental learning methods, yet both rely on task-ids, and their evaluation performance in the class-IL setting is suboptimal \cite{zhang2022cglb}.

\subsection{Implementation Details}
\label{app:Implementation details}

For GNN backbones, we set their depth to be 2 layers, adopting the implementations from the PyTorch Geometric Library\footnote{\url{https://github.com/pyg-team/pytorch_geometric}} in all experiments. 
The hidden size is varied within the set $\{16, 32, 128\}$ to accommodate different datasets.
We utilize the Adam optimizer with the learning rate is set to 0.01, and weight decay is set to $5\times10^{-4}$
In the pre-training phase, the training process will run for 200 epochs, and in the NCD-training phase, it will run for 600 epochs. 
Furthermore, early stopping is implemented to prevent model degeneration.
The parameters of the above baseline methods are set as the suggested value in their papers or carefully tuned for fairness.
We introduce an additional hyper-parameter $\lambda$, which is set within the range of 0.4 to 1, to balance the performance of the model and avoid learning new categories too weakly or forgetting old categories too seriously. Other coefficients are adjusted for different datasets, including $\beta_1 \in [0.05, 0.1]$, $\beta_2 \in [4, 5]$ and $\eta \in [0.2, 1]$.

\subsection{Additional Experiments and Analysis}
\subsubsection{Analysis with Average Accuracy and Forgetting} \label{app:aaaf}
We additionally employ average accuracy (AA) and average forgetting (AF) \cite{zhang2022cglb} to evaluate and compare the performance of all methods. These metrics serve to characterize the average performance of each model in the context of learning new classes and the resistance to forgetting, respectively.
Considering $n$ learning phases, let the comprehensive performance matrix be denoted as $M \in \mathbb{R}^{n \times n}$. For phase $i$, the $j$-th ($j \le i$) element in the $i$-th row of $M$ corresponds to the accuracy of the current model on the learnt old classes in phase $j$. The AA for phase $i$ is calculated through $\frac{1}{i} \sum\nolimits_{j=1}^{i}M_{ij}$, while the AF is assessed by $\frac{1}{i-1} \sum\nolimits_{j=1}^{i-1}{M_{ij}-M_{jj}}$. Notably, as there is no forgetting for the initial task, its AF is inherently 0.

\begin{figure*}[!t]
\centering
\subfloat[SWORD]{
\includegraphics[width=.23\textwidth]{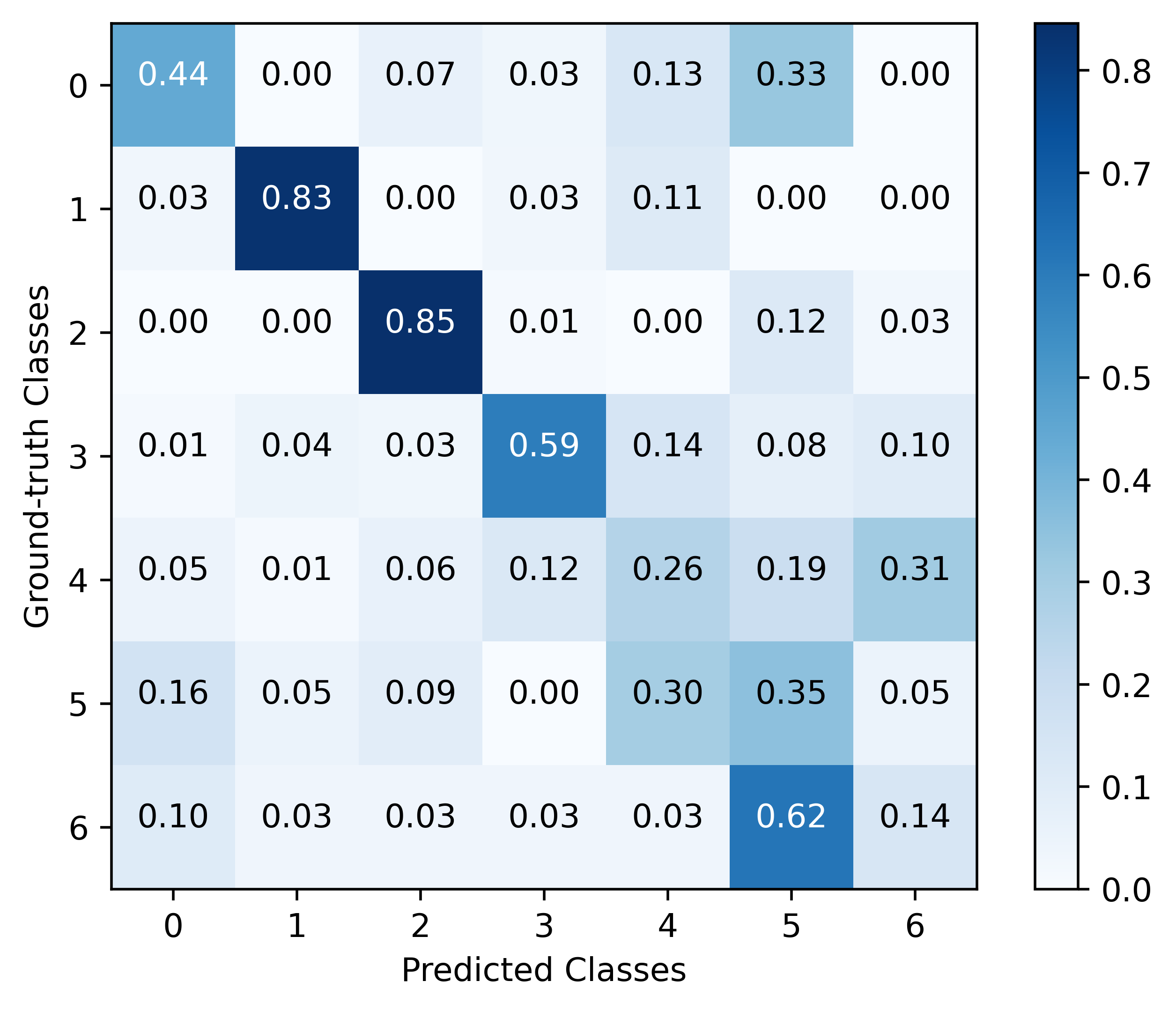}} 
\subfloat[ResTune]{
\includegraphics[width=.23\textwidth]{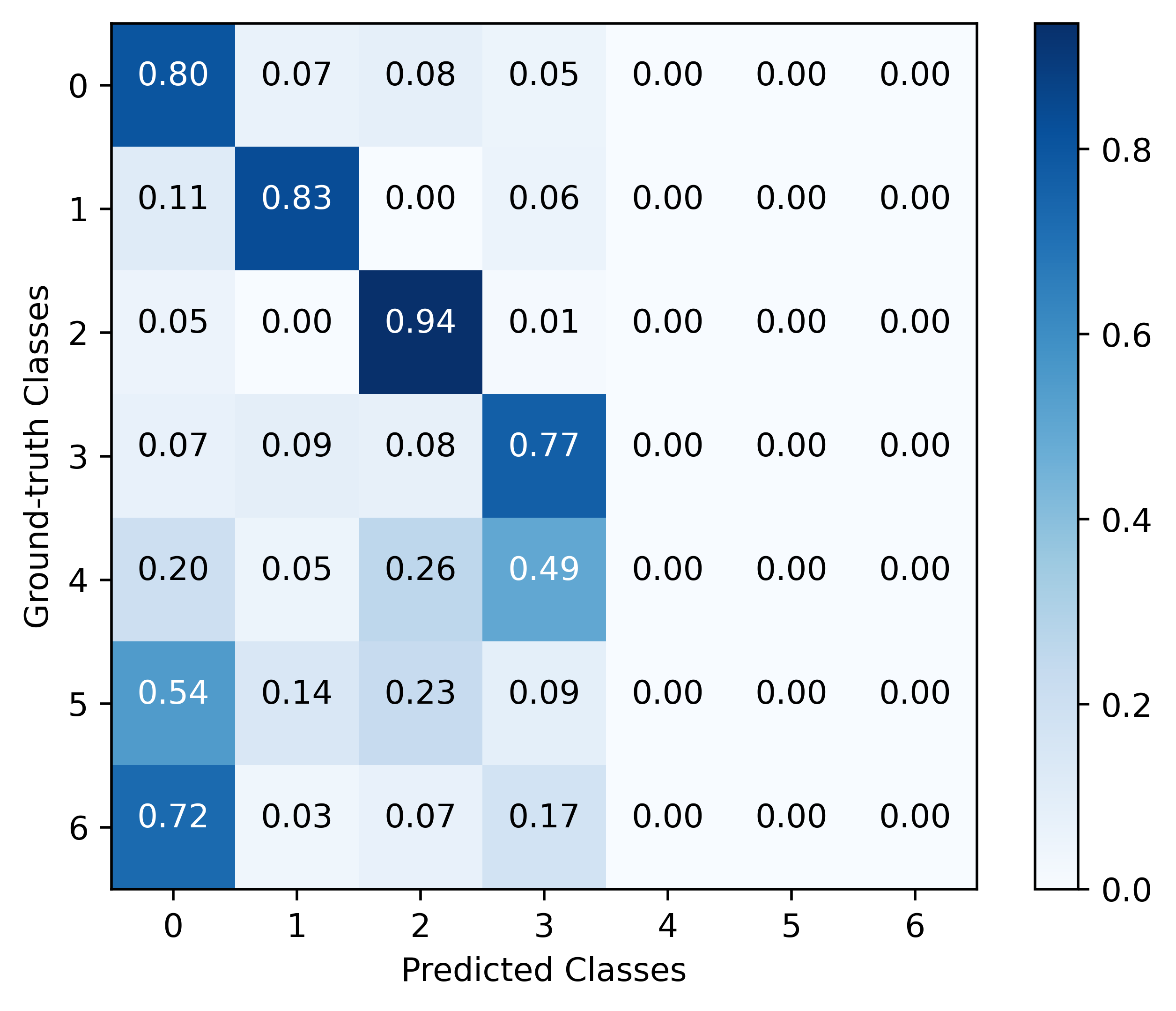}} 
\subfloat[DTC]{
\includegraphics[width=.23\textwidth]{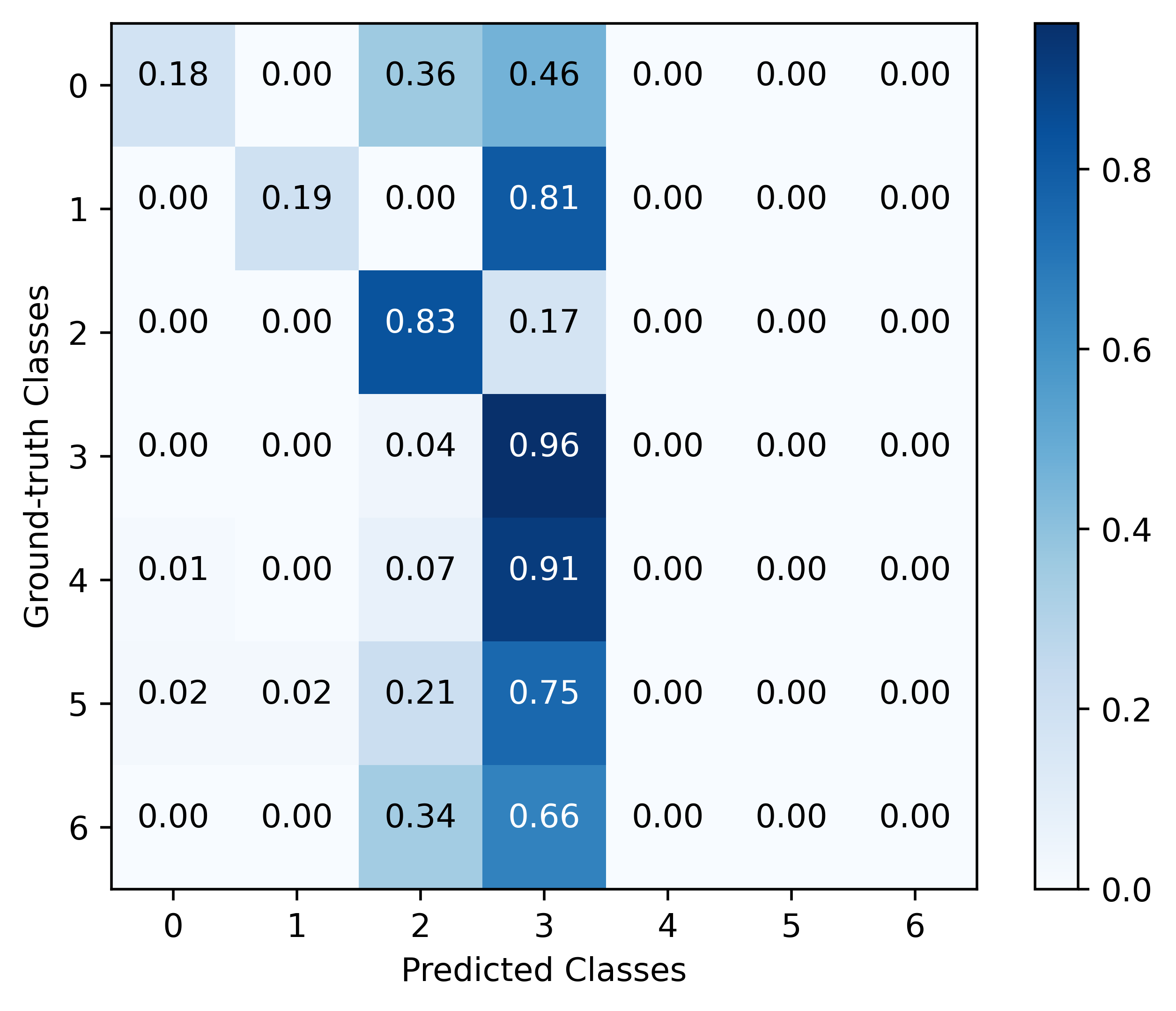}}
\subfloat[ER-GNN]{
\includegraphics[width=.23\textwidth]{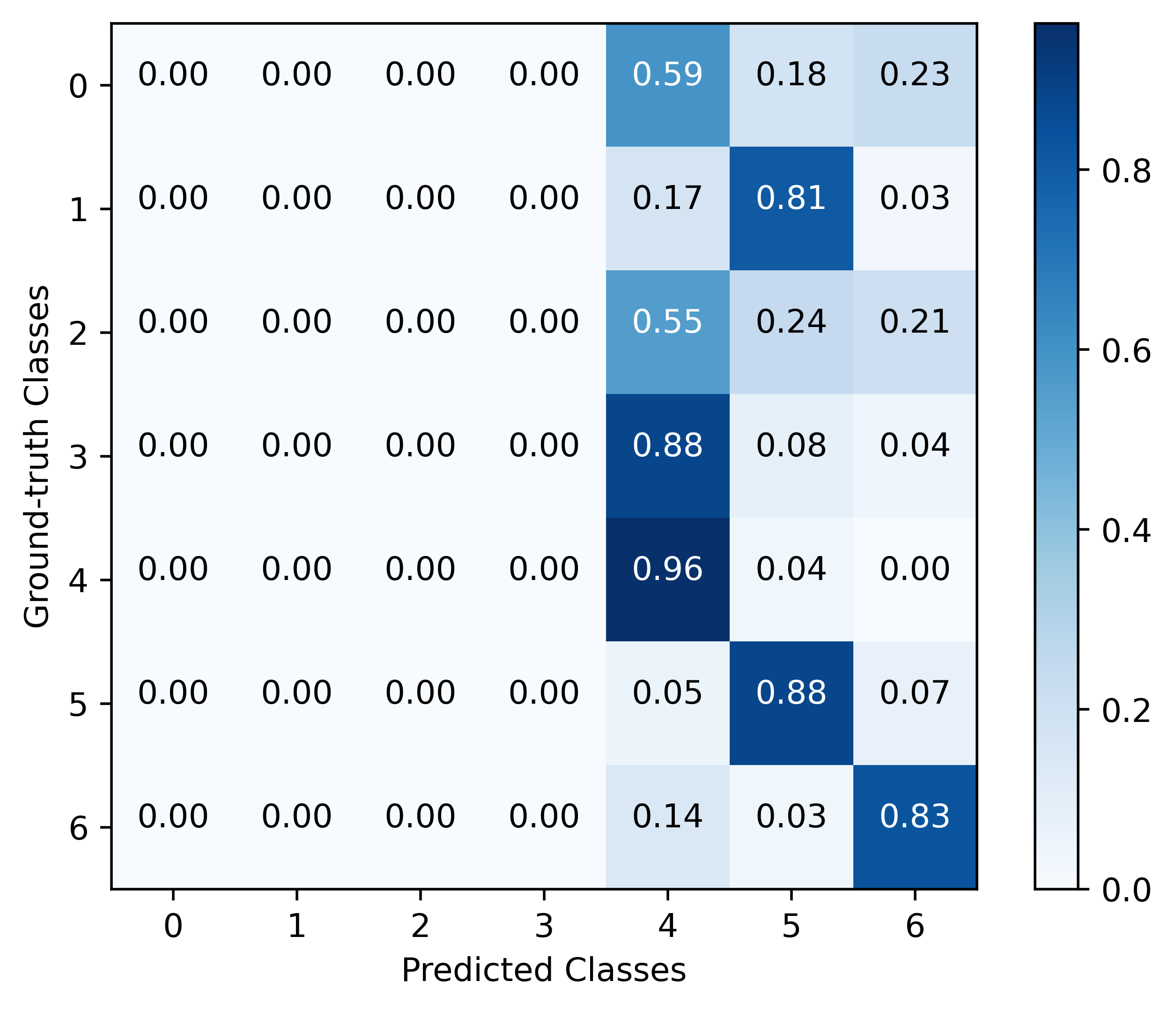}} \\
\subfloat[CPCA]{
\includegraphics[width=.23\textwidth]{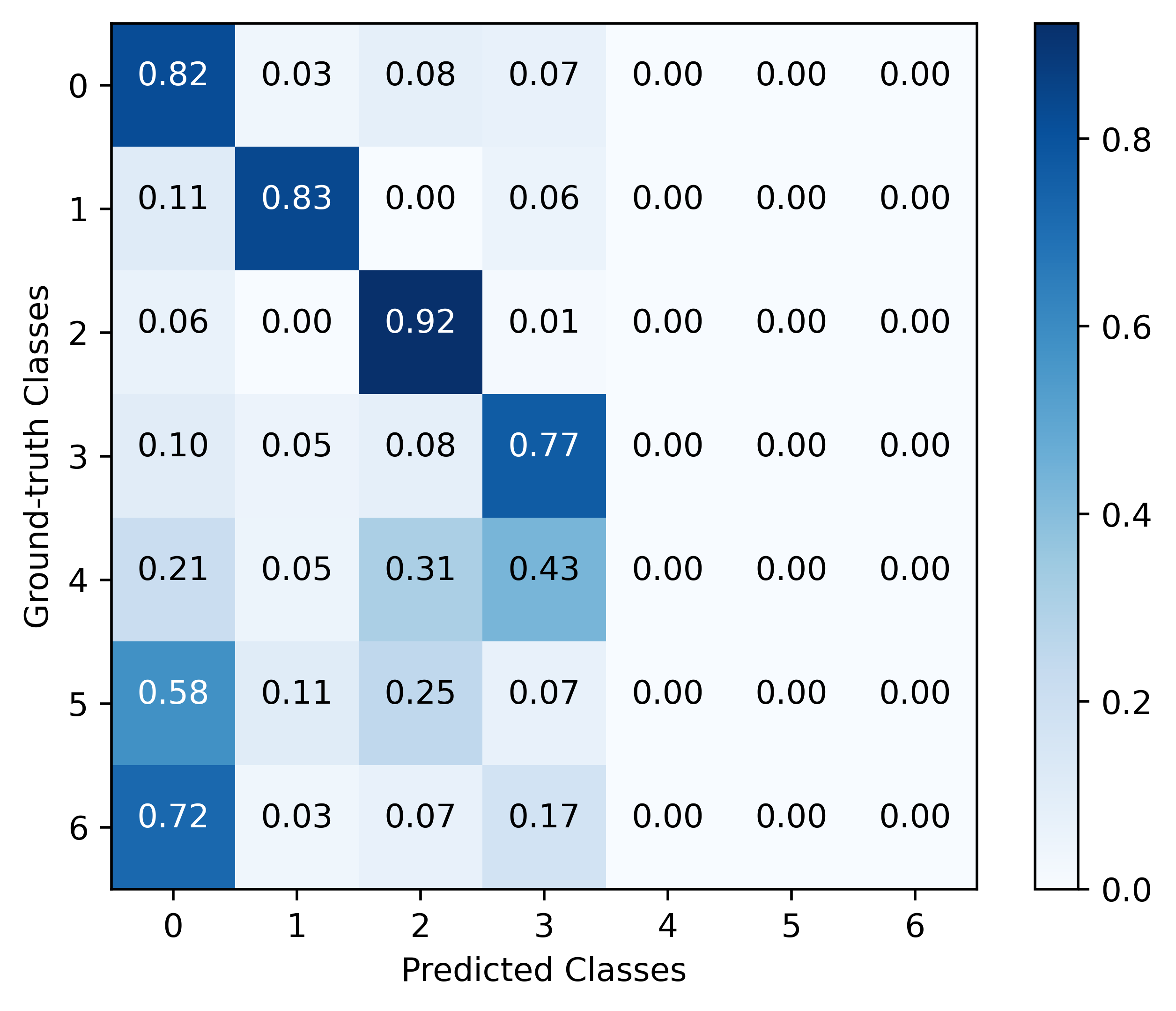}}
\subfloat[SWORD w/o ST]{
\includegraphics[width=.23\textwidth]{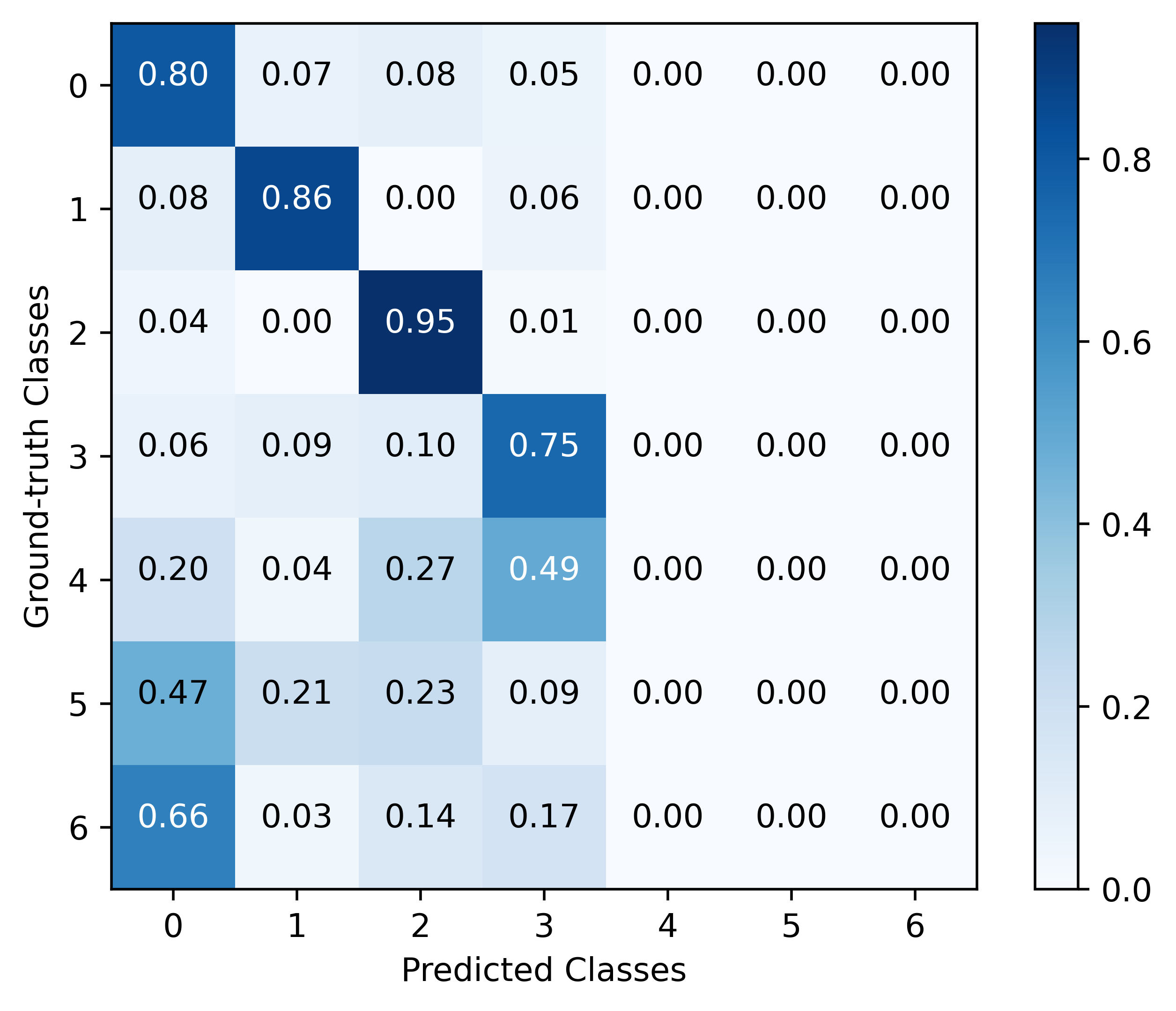}} 
\subfloat[SWORD w/o FR]{
\includegraphics[width=.23\textwidth]{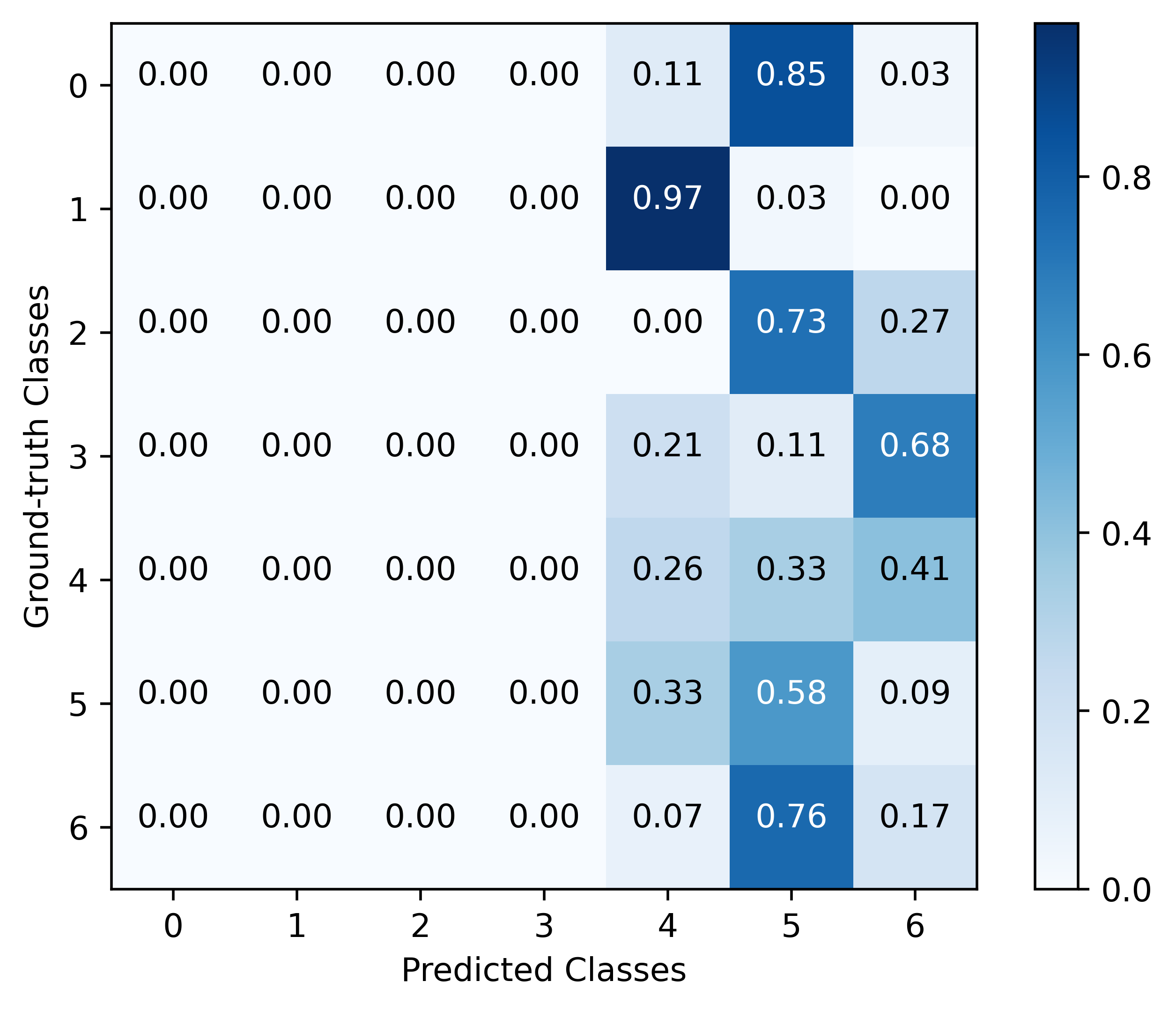}}
\subfloat[SWORD w/o ST,FD,FR]{
\includegraphics[width=.23\textwidth]{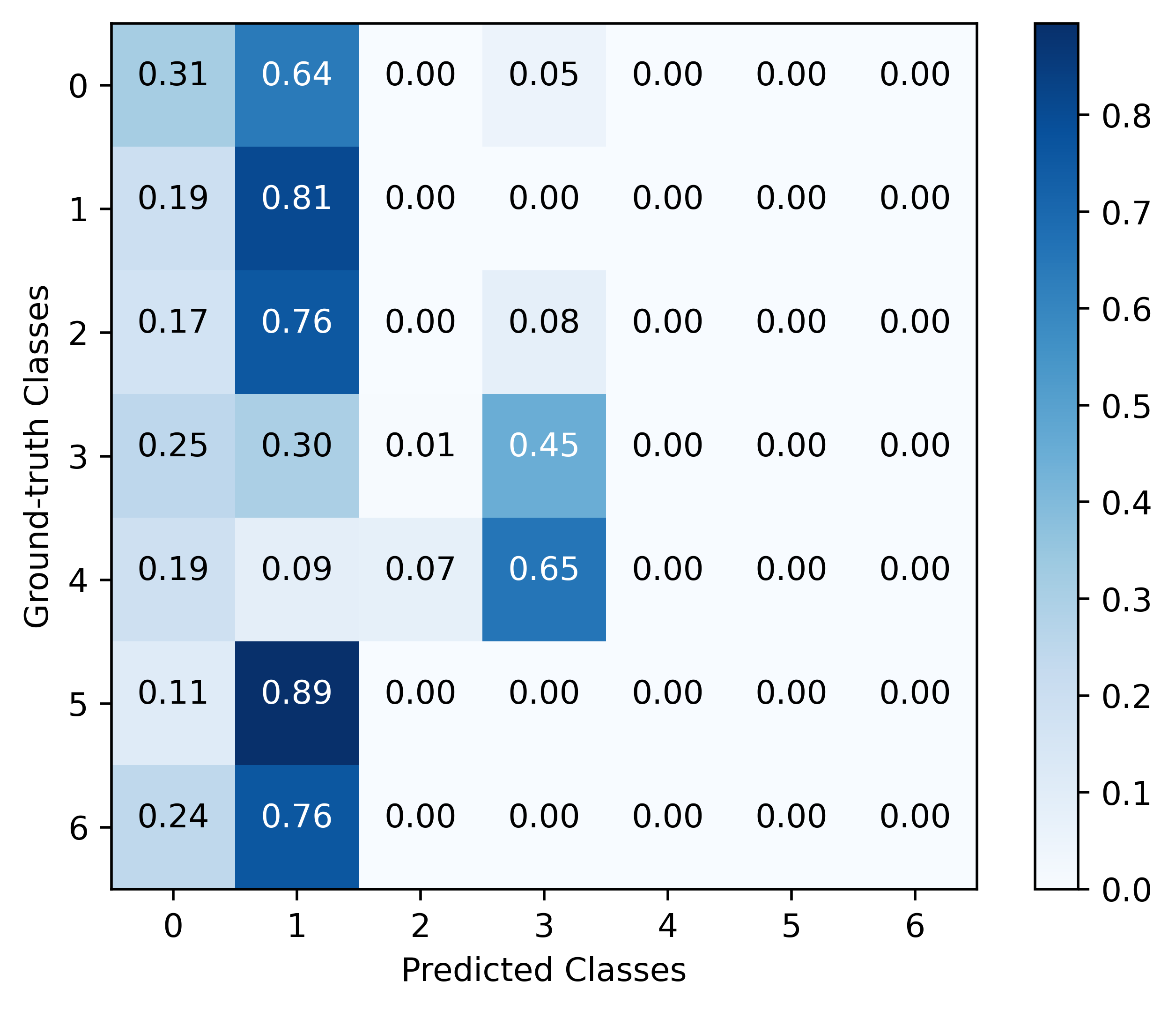}}
  \caption{Comparisons of confusion matrix of different methods.}
  \label{fig:confusion-matrix}
\end{figure*}

\begin{figure*}[!t]
\centering
\subfloat[Original Graph]{
\includegraphics[width=.15\textwidth]{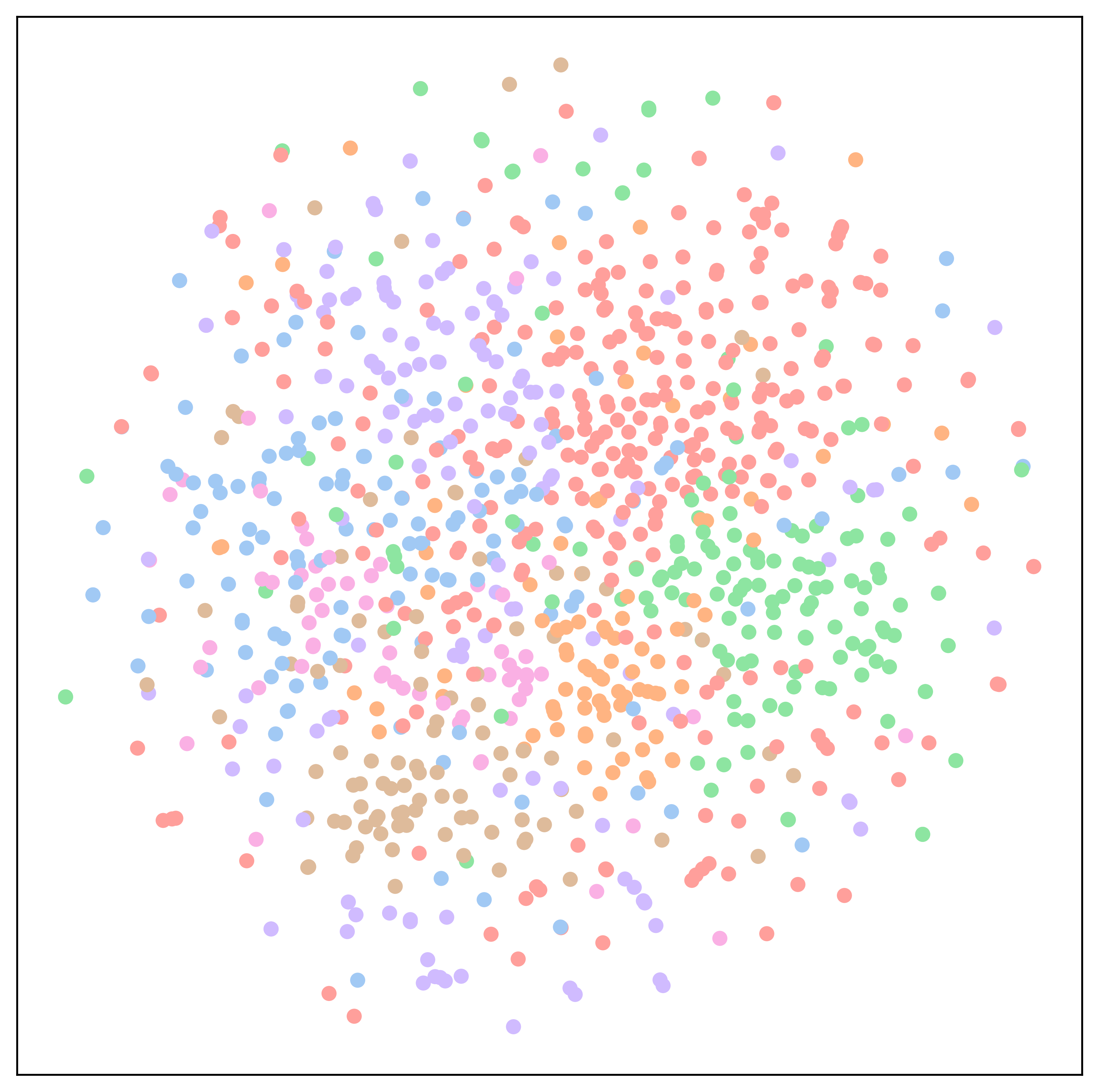}}
\subfloat[SWORD]{
\includegraphics[width=.15\textwidth]{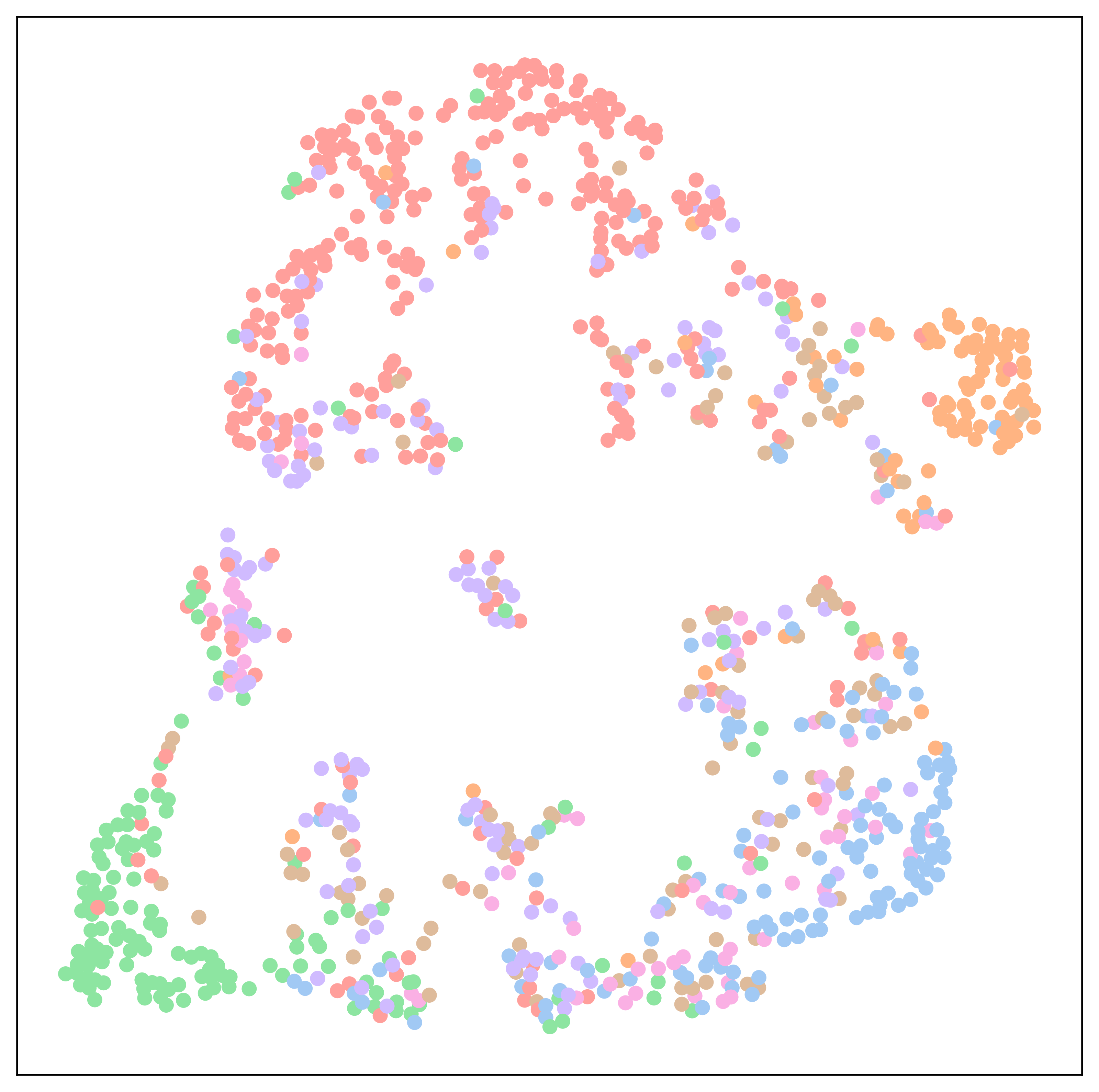}}
\subfloat[AutoNovel]{
\includegraphics[width=.15\textwidth]{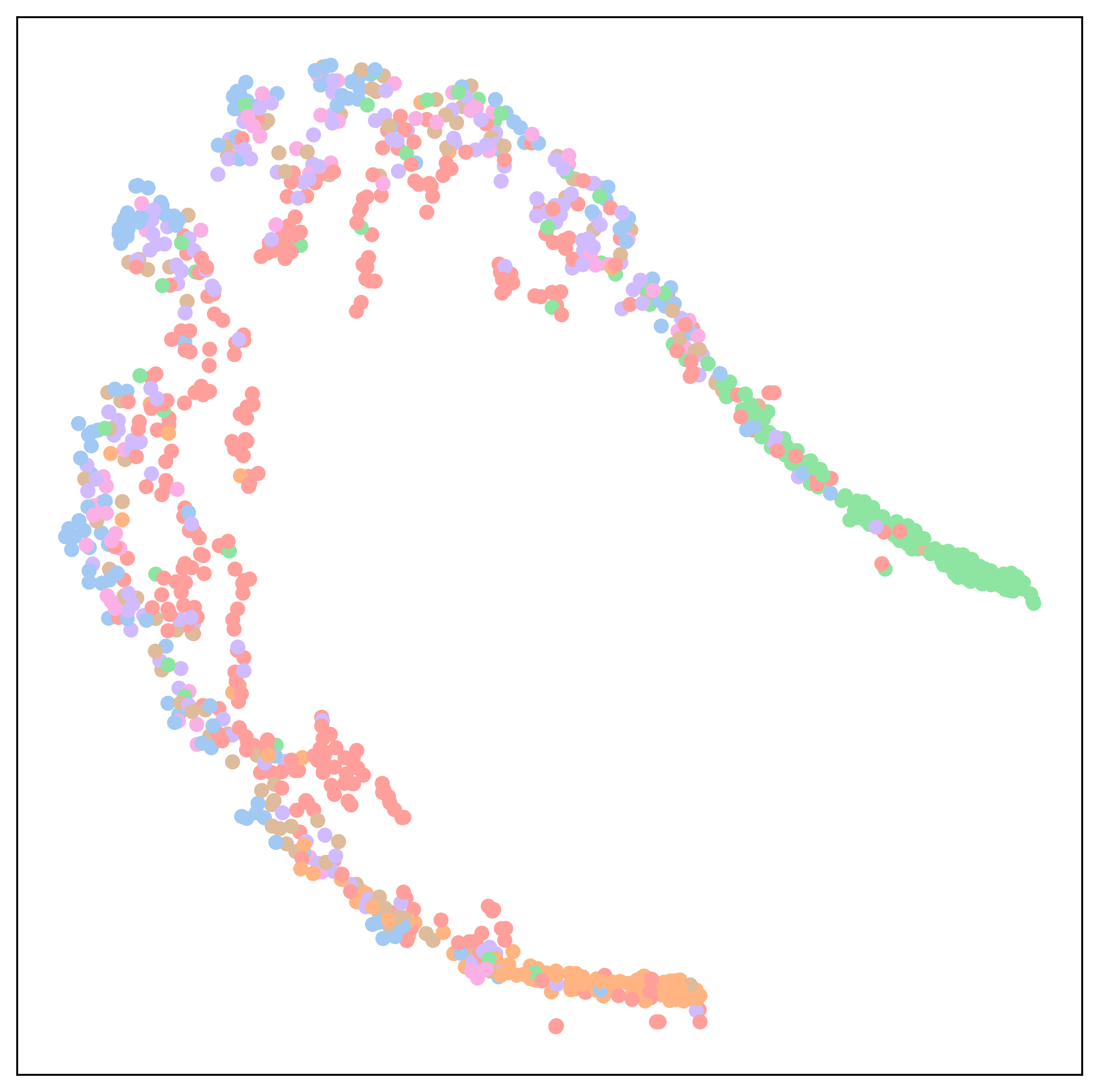}}
\subfloat[NCL]{
\includegraphics[width=.15\textwidth]{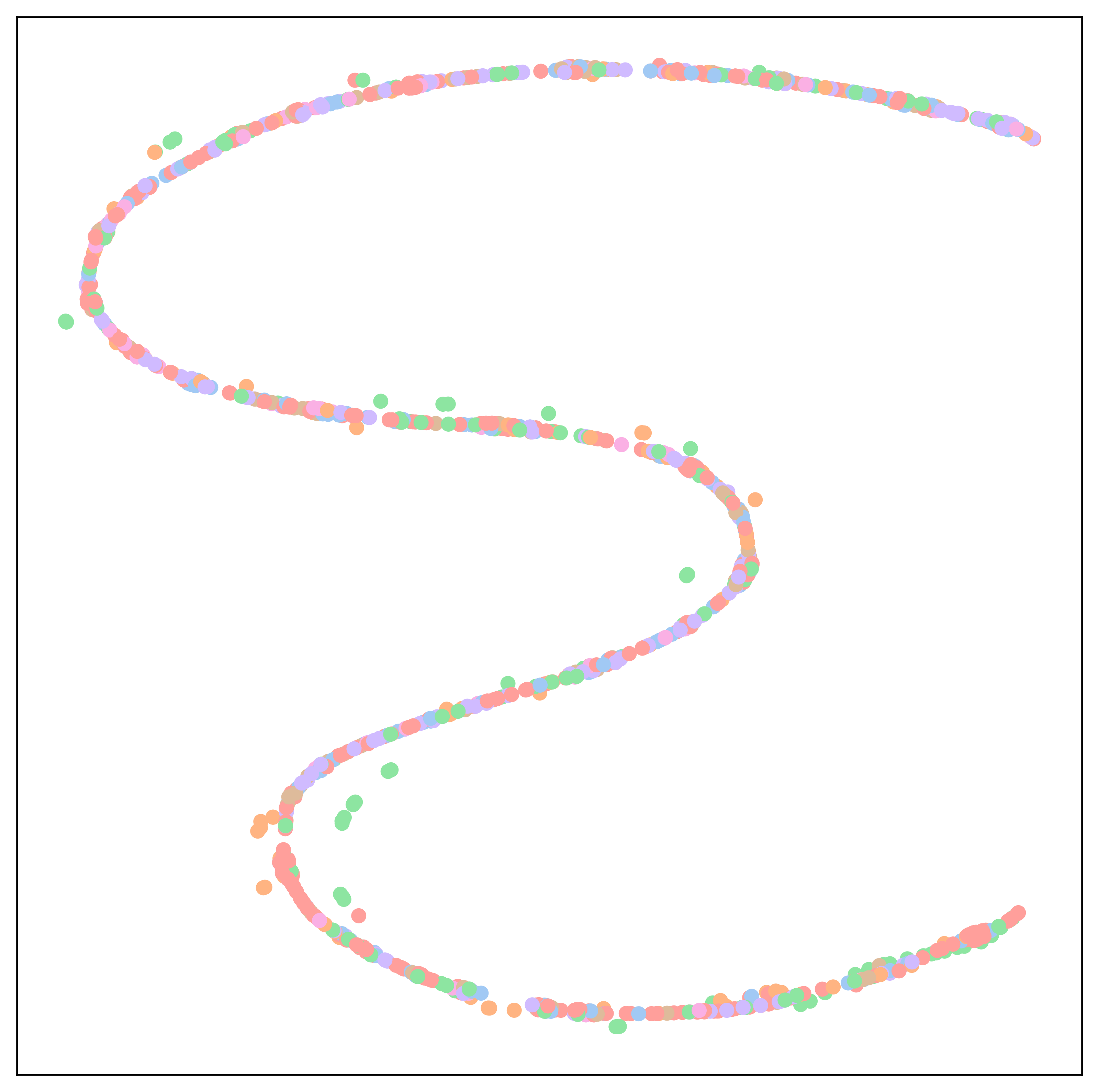}}
\subfloat[GEM]{
\includegraphics[width=.15\textwidth]{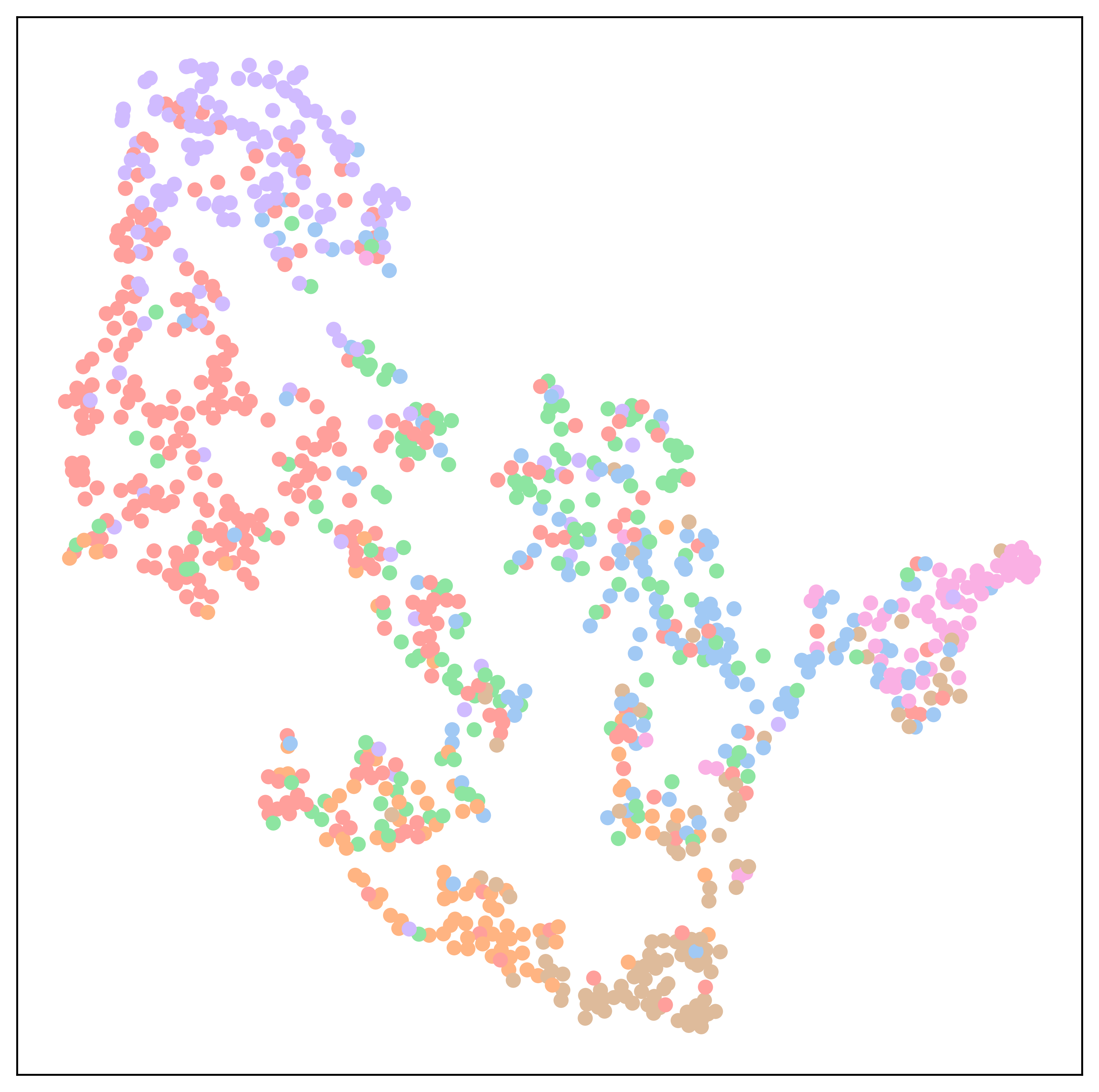}}
\subfloat[ER-GNN]{
\includegraphics[width=.15\textwidth]{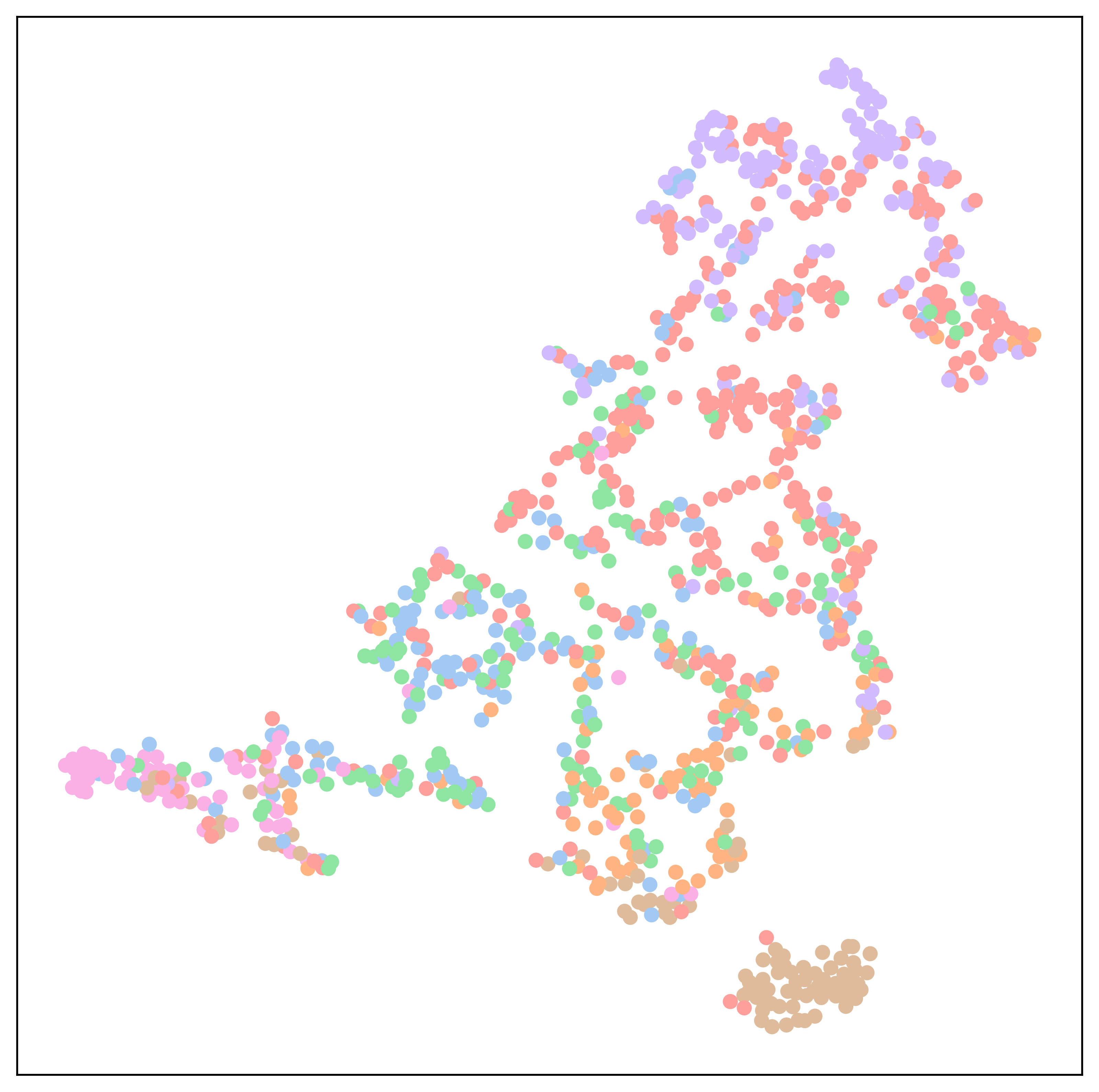}}
  \caption{T-SNE embedding visualization.}
  \label{fig:visual}
\end{figure*}

As shown in Table \ref{table:aaaf}, our SWORD exhibits a comparatively high performance on AA, primarily attributed to its ability to effectively balance the classification performance of both new and old category nodes. The elevated performance in AF of SWORD indicates a reduced level of forgetting of old categories after learning new ones.
While several NCD methods also demonstrate commendable performance on AF, our analysis from Table \ref{table:comp} reveals that this comes at the cost of the model's inability to effectively learn new categories, resulting in poor AA performance.
It is noteworthy that, despite TWP, ER-GNN, and GEM achieving respectable AA performance on individual datasets (eg., Cora, Citeseer), their significant forgetting shown in AF is apparent, stemming from reliance on labeled data for old categories during the training stage. 
This underscores the importance of evaluating model performance in the NC-NCD setting, where metrics simultaneously assess the model's performance on new, old, and all categories. Such evaluation protocols prove more effective in distinguishing performance differences and actual capabilities among various baseline models in NCD tasks.

\subsubsection{Analysis with confusion matrix} \label{app:matrix}
To further examine the performance of the aforementioned methods in classifying nodes during the inference phase, we present the corresponding confusion matrix in Figure \ref{fig:confusion-matrix}.
Upon reviewing Table \ref{table:comp}, we observe that although ResTune and DTC exhibit high accuracy in classifying old categories and all categories after learning new nodes, they struggle to accurately classify new nodes.
These confusion matrices illustrate that these baseline methods tend to assign a majority of the test samples to the old categories. It is through this approach that the impression of better performance on old categories, as well as all categories, is maintained by these baseline methods.

As mentioned above, ResTune, as a task-agnostic approach, aims to address the forgetting problem in NCD tasks. It outperforms other NCD methods in classifying old categories. However, the confusion matrix in Figure \ref{fig:confusion-matrix}(b) reveals that ResTune fails to learn the new category nodes and still incorrectly distinguishes the old and new category nodes in a task-agnostic manner.
Indeed, the inferior performance on new categories is not attributed to the inability to learn new nodes. In fact, these baseline methods are only capable of distinguishing nodes that appear in task $\mathcal{T}^u$ when the specific classifier $h^u$ is used with the task-id specified.
However, this is not aligned with the actual setting.

The aforementioned observations highlight a flaw in the current NCD methods, including ResTune, in terms of achieving a balanced performance between old and new categories. Moreover, this finding emphasizes the practical significance of the NC-NCD setting, which enables a more accurate assessment of the method's effectiveness with our new evaluation protocol.
The confusion matrix in Figure \ref{fig:confusion-matrix}(a) demonstrates the proper and effectiveness of SWORD, which correctly classifies nodes in all known categories and consistently performs well in all test datasets.

\subsubsection{Visualization}  \label{app:tsne}
We visualize the Cora dataset's original graph, as well as the embedding of graphs learned by SWORD and other baseline methods (i.e., AutoNovel, NCL, GEM, ER-GNN) via t-SNE. 
As shown in Figure \ref{fig:visual}, each color corresponds to a specific category.
It's evident that SWORD is able to cluster all category nodes better, while the other baseline methods are not as capable of distinguishing between the individual category nodes.

\subsubsection{Impact of the Number of Backbone's Layers}  \label{app:layers}

\begin{figure}[!t]
\centering
\subfloat[Layers of GCN]{
\includegraphics[width=.16\textwidth]{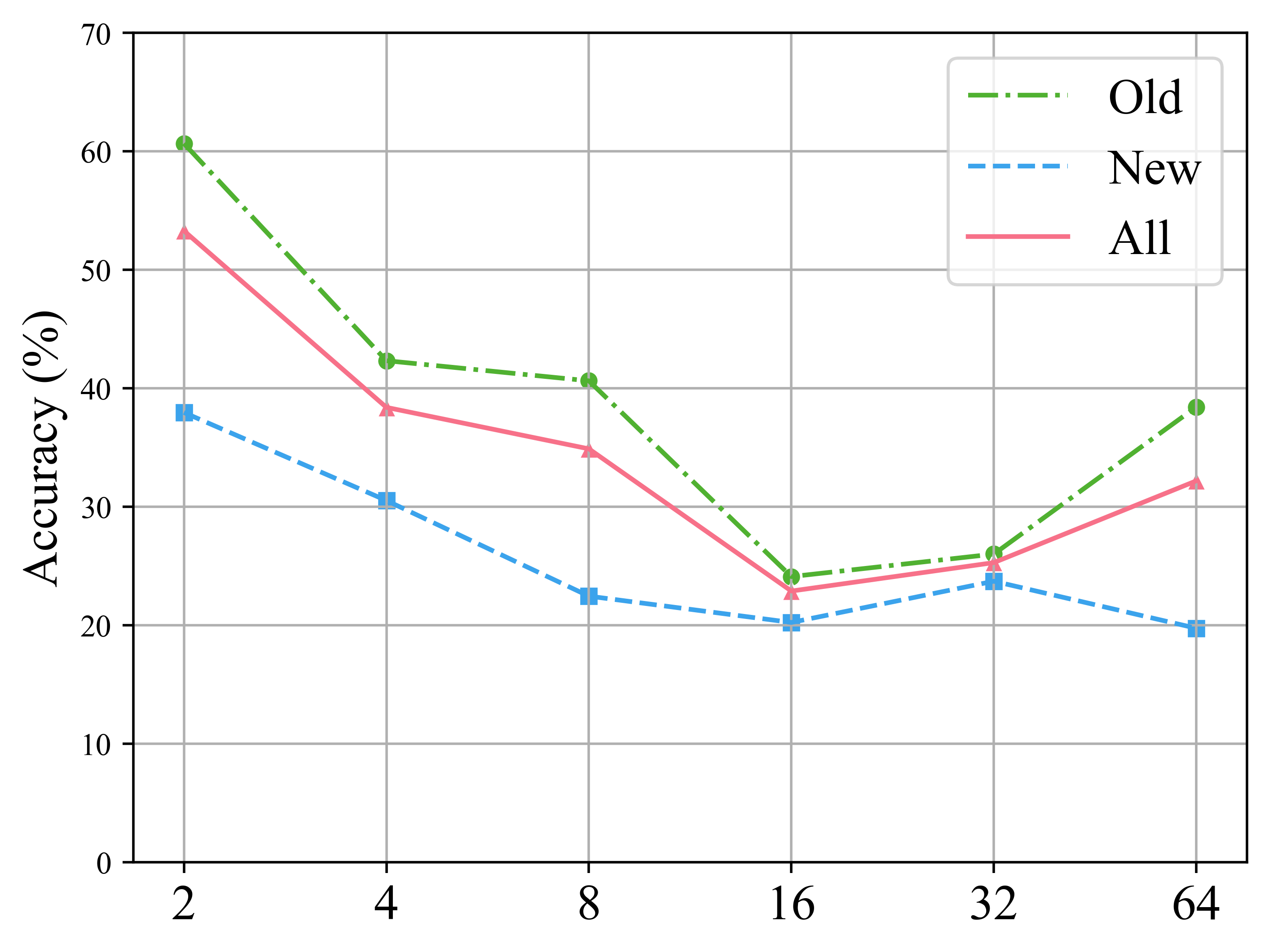}}
\subfloat[Layers of GAT]{
\includegraphics[width=.16\textwidth]{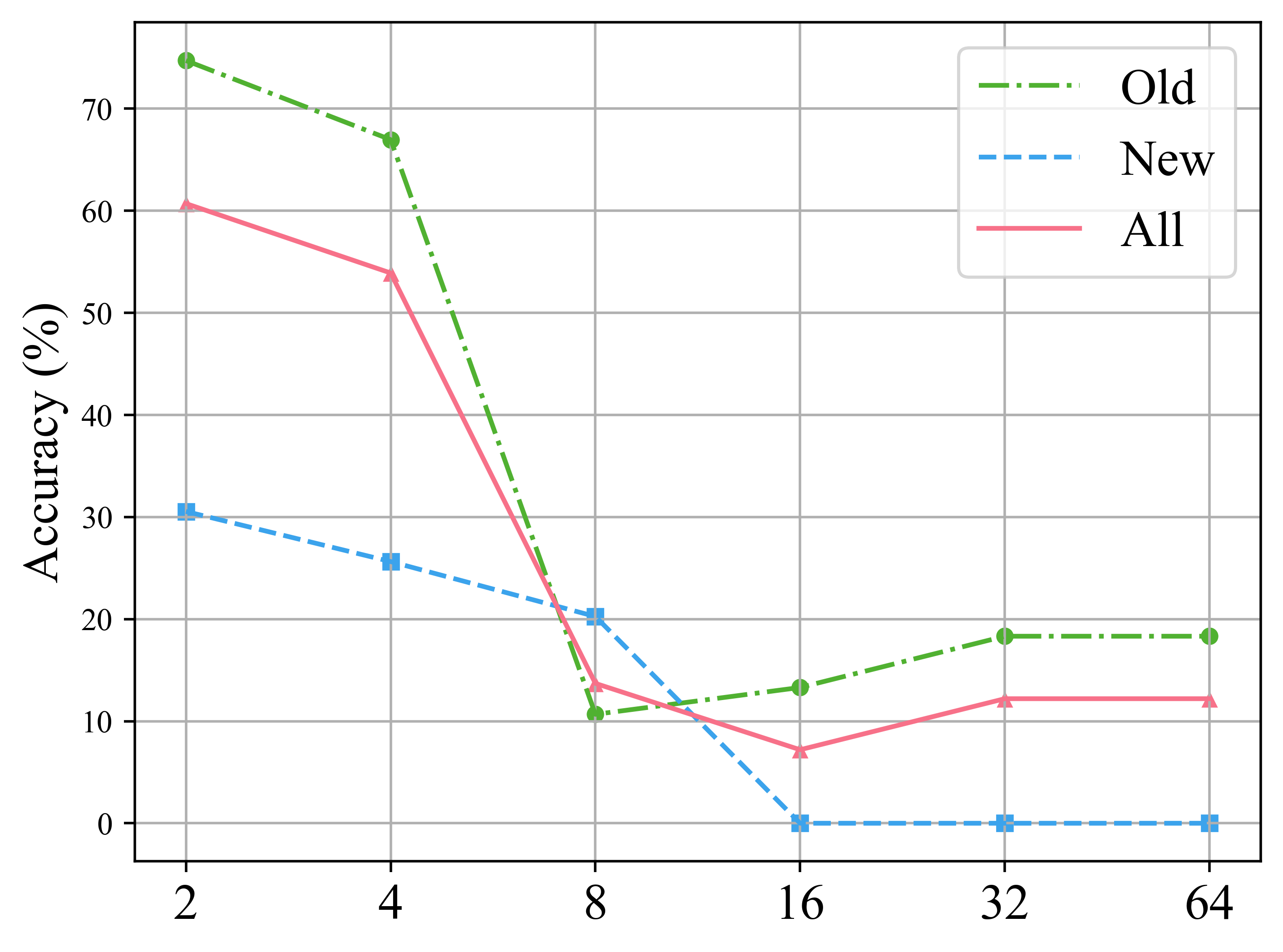}}
\subfloat[Layers of GraphSAGE]{
\includegraphics[width=.16\textwidth]{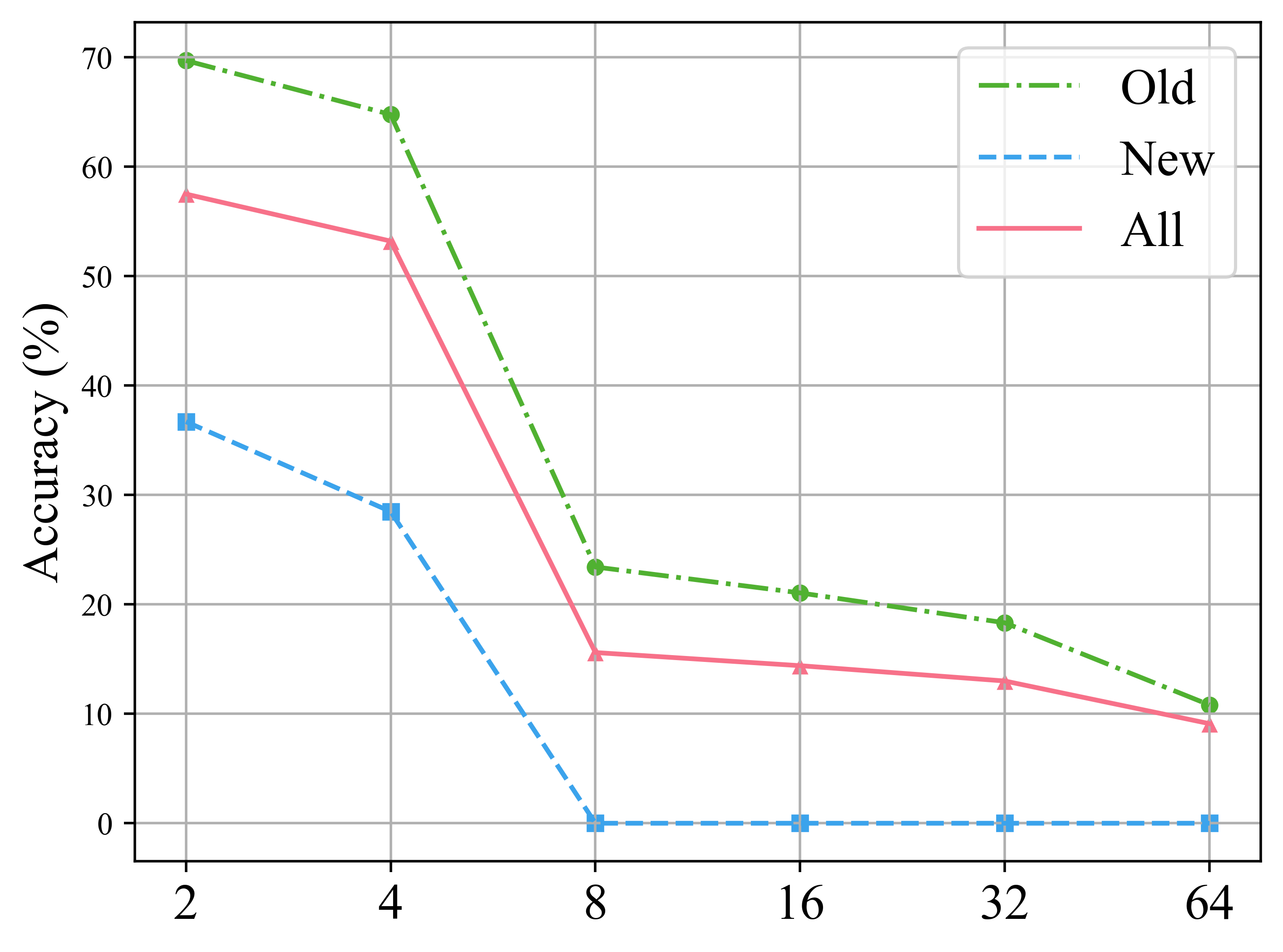}}
\caption{Classification performance with varying number of layers.}
  \label{fig:layer}
\end{figure}

As shown in Figure \ref{fig:layer}, we conduct experiments to analyze the impact of the number of layers $\in \{2, 4, 8, 16, 32, 64\}$ in the GNN backbones on classification performance. 
It can be observed that GNNs with 2 layers achieve the best results in the NC-NCD task. As the number of layers increases, the classification performance of all GNN backbones tends to decrease.
In particular, GAT and GraphSAGE with high layer counts not only suffer from catastrophic forgetting of old categories but also result in the complete loss of the ability to learn new categories. Based on these results, we conclude that GNNs with fewer layers are more suitable for effectively classifying both old and new category nodes in the NC-NCD setting.
This observation is likely linked to the over-smoothing phenomenon that manifests as the network depth increases.

\end{document}